\documentclass[preprint,10pt,journal,compsoc]{IEEEtran}                
\ifCLASSINFOpdf
  \pdfoutput=1\relax                   
  \usepackage[]{graphicx}                
  \DeclareGraphicsExtensions{.pdf,.png,.jpg,.jpeg} 
\else
  \usepackage{graphicx}                
  \DeclareGraphicsExtensions{.eps}     
\fi%

\graphicspath{{figures/}{pictures/}{images/}{./}} 

\usepackage{microtype}                 
\PassOptionsToPackage{warn}{textcomp}  
\usepackage{textcomp}                  
\usepackage{amsmath}
\usepackage{amssymb}
\usepackage{bm}
\usepackage{mathptmx}                  
\usepackage{times}                     

\ifCLASSOPTIONcompsoc
  \usepackage[nocompress]{cite}
\else
  \usepackage{cite}
\fi

\usepackage{tabu}                      
\usepackage{booktabs}                  

\usepackage{hyperref}
\usepackage{multirow}
\usepackage{todonotes}
\usepackage{subcaption}
\usepackage{flushend}
\newcommand{\etal}{~\textit{et~al}. }
\DeclareMathOperator*{\argmax}{arg\,max} 

\title{Monocular Depth Decomposition\\ of Semi-Transparent Volume Renderings}

\author{Dominik Engel, Sebastian Hartwig and Timo Ropinski

\setcounter{figure}{0}
\begin{center}
  \centering
  \includegraphics[width=\linewidth]{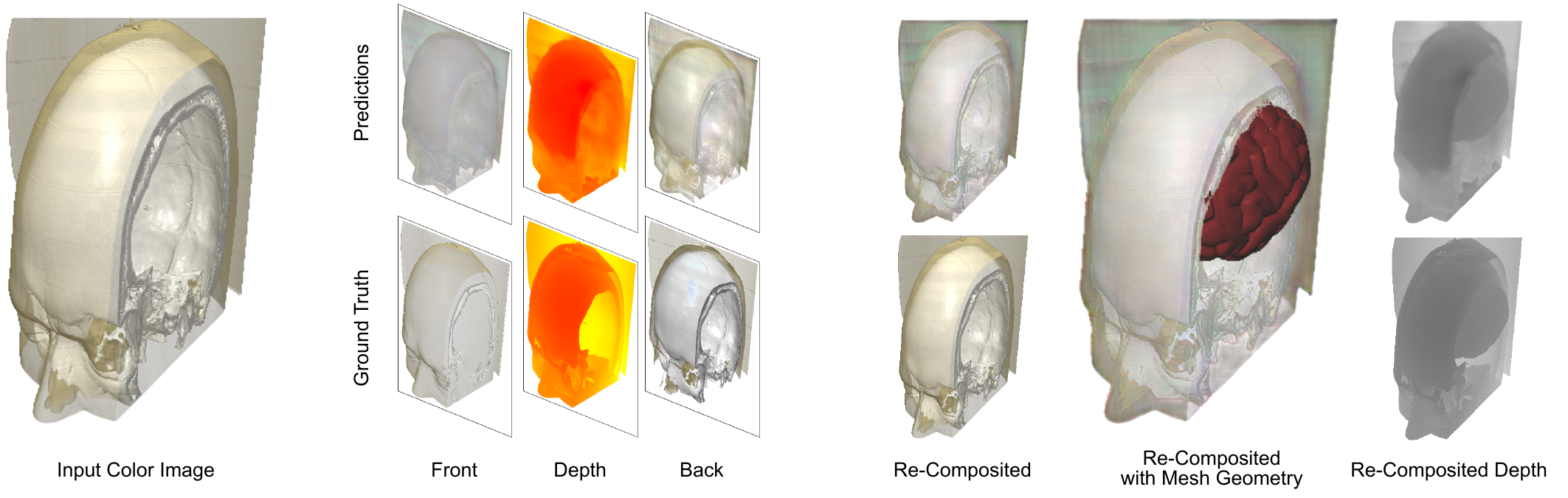} 
  \captionof{figure}{\textbf{Illustration of our approach.} Based on a single RGB \emph{Input Color Image} (\emph{left}), our neural model predicts a layered image representation (\emph{middle}), consisting of a \emph{Front} layer containing semi-transparent structures in front of the surface of interest, a \emph{Depth} layer containing the depth of the surface of interest, and a \emph{Back} layer containing structures behind the surface of interest. These layers can be re-composited to obtain the original image plus alpha, whereby alpha was not part of the input. The predicted layers allow for several visualization-based image modifications, such as for instance the integration of mesh geometries into the input image, as re-composited depth values can be obtained (\emph{right}).}
  \label{fig:teaser}
  \vspace{-0.5cm}
\end{center}%

\IEEEcompsocitemizethanks{\IEEEcompsocthanksitem Authors are with the Visual Computing Group, Ulm University.\protect\\
\IEEEcompsocthanksitem E-mail: \{dominik.engel | sebastian.hartwig | timo.ropinski \}@uni-ulm.de
}
\thanks{Manuscript received August 25, 2022; revised 3 February 2023; accepted 9 February 2023. \textcopyright 2023 IEEE \url{https://doi.org/10.1109/TVCG.2023.3245305}}} 

\IEEEtitleabstractindextext{%
\begin{abstract}
Neural networks have shown great success in extracting geometric information from color images. Especially, monocular depth estimation networks are increasingly reliable in real-world scenes. In this work we investigate the applicability of such monocular depth estimation networks to semi-transparent volume rendered images. As \emph{depth} is notoriously difficult to define in a volumetric scene without clearly defined surfaces, we consider different depth computations that have emerged in practice, and compare state-of-the-art monocular depth estimation approaches for these different interpretations during an evaluation considering different degrees of opacity in the renderings. Additionally, we investigate how these networks can be extended to further obtain color and opacity information, in order to create a layered representation of the scene based on a single color image. This layered representation consists of spatially separated semi-transparent intervals that composite to the original input rendering. In our experiments we show that existing approaches to monocular depth estimation can be adapted to perform well on semi-transparent volume renderings, which has several applications in the area of scientific visualization, like re-composition with additional objects and labels or additional shading.
\end{abstract} 

\begin{IEEEkeywords}
Volume rendering, depth compositing, monocular depth estimation.
\end{IEEEkeywords}
}

\IEEEaftertitletext{\vspace{-0.3cm}}

\begin{document}


\maketitle
\IEEEraisesectionheading{\section{Introduction}\label{sec:introduction}}
\IEEEPARstart{V}{isualization} of volumetric data is a common necessity in many applied sciences, to explore new phenomena, understand the acquired measures or to make diagnoses based on them. Methods like direct volume rendering (DVR) are commonly used in practice nowadays, and countless visualizations have been created using DVR or similar techniques. In general, DVR is the preferred method to visualize 3D data using semi-transparent structures, allowing for inspection of multiple structures at once that would normally occlude each other. However, such volume renderings usually have to be created with great care and often require a significant amount of manual labor to find good rendering parameters, such as the transfer function (TF), which assigns optical properties to the data. In practice, parameters like the TF, that are essential in order to reproduce such a visualization, are often not shared, or even the underlying original data is unavailable, which makes volume renderings impossible to fully reproduce or change in a meaningful way. While differentiable volume rendering~\cite{weiss2021differentiable} allows for reconstruction of such parameters, it usually requires access to the original underlying data and/or a subset of the parameters in question, making modifications unfeasible from just a single image without any additional information.

In this work, we investigate the abilities of neural networks to extract additional relevant information from single RGB images of such visualizations, with the goal of making these rendered images re-usable through composition or re-lighting. A perfect reproduction of such a rendered scene would reconstruct the original volume data, the transfer function and camera parameters, which is currently not feasible from just a single color image. As a more feasible goal we aim to reproduce a layered representation, which models the volumetric scene in a view-dependent manner using pre-integrated intervals together with their depth along the viewing rays. This type of representation is known from prior work in remote rendering~\cite{shade1998layered} and allows for various modifications of the produced rendering, like composition with additional objects, re-lighting or viewpoint changes. As basis for our approach, we first investigate existing approaches for monocular depth estimation and adapt them to semi-transparent scenes, in order to predict the depth of relevant structures in the image. For this, a less ambiguous definition of \emph{depth} is required, as semi-transparent renderings do not have a single depth value as opaque scenes, but rather have multiple relevant structures lying behind each other on a view ray. Looking at prior work, there are some attempts to find the \emph{most relevant} of these structures in the context of picking (e.g.,~\cite{wiebel2012wysiwyp,argelaguet2013survey}). We employ these techniques to extract depth information for surfaces of interest, and refer to these techniques as \emph{depth techniques} in the remainder of this paper. Based on this, we extend the depth estimators to additionally predict the color and opacity in \emph{front} and in the \emph{back} of the predicted depth, resulting in a layered representation that allows for modification of the scene.

Throughout this paper, we compare the performance of multiple state-of-the-art (SotA) monocular depth estimation networks for different \emph{depth techniques} and different \emph{degrees of opacity} in the rendered images. To our knowledge, this is the first study considering monocular depth estimation on volume rendered images. Based on our findings, we modify the SotA models to not only predict the depth of the surface of interest, but also the color and opacity of the semi-transparent structures in front, as well as the color behind this surface of interest (see Fig.~\ref{fig:teaser}), corresponding to a 2-layer representation. Based on the obtained layers, which can be predicted from a single RGB-only volume rendered image, layer-wise re-compositing of the original rendering becomes possible, such that for instance additional meshes can be faithfully integrated into the volume rendering. All without any knowledge about the underlying volume data, transfer function or any other rendering parameter. 

To achieve our goals, we make the following contributions within this paper:
\begin{itemize}
    \item We evaluate SotA monocular depth estimators on volume rendered images subject to different amounts of semi-transparency.
    \item We extend the best performing model to predict a layered image representation, containing the depth of a visually dominant surface, as well as color and alpha layers in \emph{front} and in the \emph{back} of these surfaces, enabling the re-use and modification of the single input color image.
    \item To enable the extraction of such layered representation, we present two novel loss functions: A compositing loss to enforce the \emph{front} and \emph{back} layers to composite to the original input image and a front-back-divergence loss to encourage the layers to properly separate different structures into different layers.
    \item We demonstrate the efficacy of our approach by compositing mesh geometries into existing volume renderings for which we only have access to RGB images. Furthermore we use the extracted layers to retrospectively apply ambient occlusion, add surface labels or change viewpoints. 
\end{itemize}
We further make our code, datasets and trained models publicly available\footnote{\url{https://github.com/xeTaiz/MonocularDepthDecomposition}} in order to enable other visualization researchers to use and further extend the presented techniques.

\section{Related Work}\label{sec:related-work}

Within this section, we discuss prior work related to our approach. We will first provide an overview about depth techniques in the context of direct volume rendering, before we briefly recap the SotA in monocular depth estimation.

\noindent\textbf{Volume rendering depth.} Traditionally, volume rendered images are generated using the emission absorption model, which is a physics-inspired model underlying the popular volume rendering integral~\cite{max1995optical}. While this approach leads to faithfully volume rendered images of the volume data at hand, the cloudy nature of the resulting imagery let researchers investigate methods to visually emphasize more well-defined and delineated objects, to support object detection and quantification. The most seminal work in this direction is most likely Levoy's early paper on gradient-based surface extraction from volume data~\cite{levoy1988display}. Kindlmann's and Durkin’s build up on this idea, to obtain gradient-based transfer functions~\cite{kindlmann1998semi}, in which they exploit the gradient magnitude at a sample in order to modulate the sample’s opacity. Further, inspired by this approach, several other transfer function approaches followed, that exploit particular properties to modulate opacity and thus obtain more surface-like representations (e.g.~\cite{kindlmann2003curvature,correa2008size,caban2008texture,prassni2010shape}). Naturally, with the opportunities to represent well delineated objects in volume rendered images, the wish to fuse volume rendered images with geometry meshes arose~\cite{kaufman1990intermixing}. While this can be directly achieved when extracting surfaces from volumes~\cite{newman2006survey}, the integration of geometry into standard ray-casting based volume renderings~\cite{kruger2003acceleration} required the generation of a meaningful depth map in order to support correct image compositing~\cite{lindholm2015hybrid}. The most straight forward approach to obtain depth values for a volume rendered image, is the use of FirstHit depth values, where the depth value represents the depth of the first non-transparent sample along a viewing ray. Needless to say, that a single depth value cannot represent the complexity of a volume rendered image containing semi-transparent structures. The same observation is true for maximum gradient magnitude based depth maps, whereby the depth value represents the depth of the sample with the maximum gradient magnitude along a view ray. Other, more advanced approaches allow for the extraction of multiple depth values along a ray. Lindholm for instance applied a-buffer based depth peeling to obtain a complex depth structure, quite faithfully representing the complex nature of the volume rendered object~\cite{lindholm2015hybrid}. Bruckner and Gr{\"o}ller presented the MIDA method, which is not particularly targeted at extracting depth values, but can obtain meaningful surface representations by exploiting the compositing gradient~\cite{bruckner2009mida}. Similarly, the \emph{what you see is what you pick} (WYSIWYP) approach allows for the extraction of several layers, originally meant to support picking of visually salient features~\cite{wiebel2012wysiwyp,stoppel2014visibility}. While we suggest using such approaches to compute depth maps during volume rendering, Stoppel and Bruckner have also shown that layers can be used to guide interaction widgets in the same context~\cite{stoppel2018smart}. Apart from these approaches, which are somehow integrated into the rendering process, also other more advanced models exist, which support the extraction of boundaries. Lindholm\etal for instance propose the improved visualization of boundary surfaces by exploiting material-specific reconstruction in the classification process~\cite{lindholm2014boundary}.

\noindent{\textbf{Layered image representations.}}
Representing volumetric scenes as a set of image layers has been explored extensively in the context of remote and distributed rendering~\cite{zellmann2012image,frey2013explorable,tikhonova2010explorable}. During rendering, different segments are identified along the rays and saved as a \emph{layered depth image}~\cite{shade1998layered}, storing the depths, colors and opacities of each segment per pixel. Such layered representations can be rendered quickly on weaker hardware, like VR/AR devices and even allows for slight viewpoint changes. Naturally a larger number of layers allows for better novel views, whereas a small number quickly results in holes in the image.
Commonly those representations are generated directly during the rendering process, when actual volumetric information is available, so that many layers can be extracted, creating a representation of the full volume, as seen from a specific viewpoint.
Our approach aims to predict such a layered representation from just a single RGB image. As we do not have full volumetric information present at inference and can only rely on what is visible within the rendering, we for now limit ourselves to predicting a 2-layer representation. Predicting such a layered representation also enables all common interactions available to such representations, like re-lighting, composition with additional geometry, etc.

\noindent{\textbf{Monocular depth estimation.}} 
After the first approaches started to leverage convolutional neural networks (CNNs) to directly estimate depth from a single input image~\cite{eigen2015predicting,eigen2014depth}, monocular depth estimation techniques developed in a rapid way. Scores on famous benchmarks such as NYU~\cite{silberman2012nyu} and KITTI~\cite{geiger2013kitti} have been surpassed multiple times. While standard CNNs reduce the spatial resolution of feature maps, other work implements multi-layer convolutional networks~\cite{laina2016deeper} and multiscale networks~\cite{eigen2015predicting,eigen2014depth,lee2019bts}. Framing depth estimation as classification task, Fu\etal introduce deep ordinal regression networks which divide depth values into discreet ordinal labels~\cite{fu2018dorn}, instead of directly regressing depth~\cite{lee2019monocular,yin2019vnl}. Maximov\etal improve depth estimation using domain invariant defocused synthetic images as supervision in order to close the reality gap~\cite{maximov2020focus}.

Segmentation masks which separate individual objects in the input image pose a strong prior to depth estimation. Wang\etal perform semantic image segmentation and apply a divide and conquer strategy to depth estimation~\cite{wang2020sdc}. Independently, for each segment, depth maps are predicted in a canonical space and then recomposed into global space. Zhu\etal use semantic segmentation masks for better depth estimation, especially along object borders~\cite{zhu2020edge}. In the work of Ramamonjisoa\etal they also focus on enhancing depth predictions around occlusion boundaries using displacement fields~\cite{ramamonjisoa2020predicting}. Each dataset comes with distinct characteristics and biases. Circumventing domain bound depth predictions, Ranftl\etal combine multiple large datasets in order to train a network that is invariant to changes in depth range and scale~\cite{ranftl2020midas}. Virtual normals have been introduced by Yin\etal enforcing high-order geometric constraints in 3D space for the depth prediction task~\cite{yin2019vnl}. Their geometric loss function projects the predicted depth values into 3D, by using a pinhole camera model, enabling to compute virtual normals, to encode geometric constraints. More recently, Zhao\etal trained a network that removes clutter and novel objects from real images, in order to improve depth prediction~\cite{zhao2020domain}. In the work of Lee\etal~\cite{lee2019bts}, they present a novel local guidance layer, which is used to compute locally-defined relative depth estimations, on different levels of resolution. While very few monocular depth estimation papers also mention transparent surfaces~\cite{choi2021selfdeco}, to our knowledge no systematic study as well as possible applications in the context of volume rendering have been investigated.

\noindent{\textbf{Deep learning in volume rendering.}}
Lastly we briefly review related works that apply deep neural nets in the context of volume rendering. While there are several works that assist a standard volume rendering approach by predicting complex lighting~\cite{engel2020dvao} or upscaling the rendered image~\cite{weiss2019volumetric}, neural nets have also been proposed for storage and compression of volumes~\cite{weiss2021fast} and even as part of the renderer itself~\cite{weiss2021deep}. Weiss\etal~\cite{weiss2020learning} propose using neural nets for adaptive sampling during rendering and with \emph{DeepDVR}~\cite{weiss2021deep} Weiss\etal introduce a framework for neural rendering in a DVR context. In the paper they replace parts of the volume rendering pipeline with neural nets to create the first neural volume renderer. For a more in-depth review of such methods, we recommend the recent survey by Wang\etal~\cite{wang2022dl4scivis}.

\section{Method}\label{sec:method}
To investigate SotA monocular depth estimation in the context of volume rendering, we have created large scale training data sets to train existing estimators. Here we first discuss these datasets, how they were generated, and how we controlled the amount of opacity in them. Furthermore, we detail how we tackle the depth ambiguity apparent in semi-transparent structures, resulting from the lack of clear surfaces. Therefore, we compare multiple depth techniques as mentioned above. We then elaborate on the differences to the common real-world datasets for monocular depth estimation and the resulting modifications of the neural models. Lastly, we detail how we extend the monocular depth estimation models to also predict color and opacity layers in addition to depth, allowing for the de-composition of the input RGB-only volume rendering to enable visualization operations, such as for instance geometry integration.

\begin{figure*}[!tb]
    \centering
    \includegraphics[width=\textwidth]{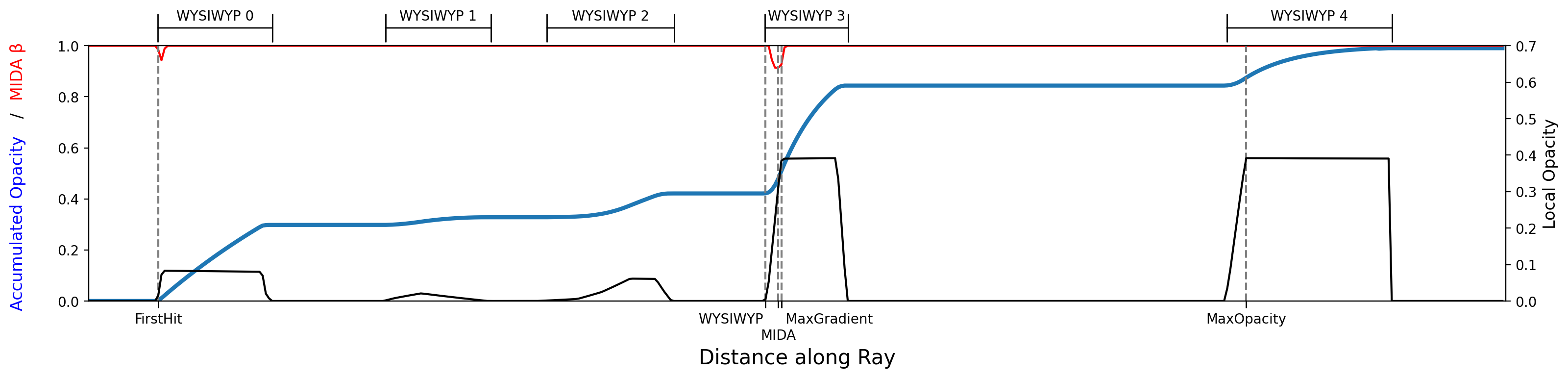}
    \caption{
    \textbf{Illustration of exploited depth techniques.} FirstHit, MaxGradient and MaxOpacity are computed from local and accumulated opacities along the ray. We acquire a depth value from the MIDA approach by determining position with minimal $\beta$. For WYSIWYP interval 3 (intervals labeled on top) is the most relevant regarding increase in accumulated opacity, and thus denotes the WYSIWYP depth in this example.}
    \label{fig:depth-technique-comparison}
    \vspace{-.45cm}
\end{figure*}
\subsection{Datasets}
\subsubsection{Dataset Properties}
The standard monocular depth estimation task requires predicting a depth map from just an RGB color image. Most existing approaches to monocular depth estimation are trained for real-world scenes, with the NYU~\cite{silberman2012nyu} and KITTI~\cite{geiger2013kitti} datasets emerging as the two most relevant ones, covering indoor and outdoor scenes. Here, we compare those \emph{standard datasets} to our requirements when applying the task to images containing semi-transparent structures. Mainly, we have identified the following three differences:

\noindent{\textbf{Depth sparsity.}}
The standard datasets generally provide quite densely labeled depth maps, with only few missing values due to reflections and occlusions in the ground truth acquisition. For the case of synthetic volume rendered images, we usually have much sparser images, as a large percentage of pixels covers background without a relevant depth. During training, this high percentage of irrelevant pixels would dominate the loss, leading to poor training signals. We tackle this problem using loss masking.

\noindent{\textbf{Absolute vs. relative depth.}}
Another relevant aspect is the depth range in the datasets. The standard datasets commonly denote their depth labels in meters, whereas synthetic renderings often do not encode real-world units at all, but rather use a rendering-centered scaling of the depth values within the view frustum. As a result, we are more interested in correctly predicting depth values relative to each other, as opposed to producing accurate absolute values. The model will learn to predict depth values that are consistent with the camera parameters seen during training, allowing us to work with relative depths within a normalized view-frustum.

\noindent{\textbf{Depth ambiguity.}}
Lastly, \emph{depth} is quite ambiguous to define in semi-transparent renderings, because there are no clear surfaces to choose, as compared to the real-world datasets. In addition to that, there may also be multiple relevant structures in front of each other without full occlusion, allowing for multiple valid choices. In this work, we first focus on predicting the \emph{most relevant} surface from a visual perspective. For this, we looked at solutions from prior work in the context of picking. A \emph{picking} layer holds information about what structure or object is "picked" (for each pixel) during a mouse interaction. Naturally, a similar ambiguity problem arises and prior work has proposed methods to decide for the \emph{most relevant} structure. We make use of this prior work and cover five different \emph{depth techniques} in Sec.~\ref{sec:depth-techniques}.

\subsubsection{Dataset Generation}
As basis for our synthetic dataset, we need a large corpus of volume data for which we can generate semi-transparent renderings automatically in order to create a sufficiently large training dataset. We chose to generate our training dataset from the CQ500 dataset by QureAI~\cite{chilamkurthy2018cq500}, which consists of 491 CT scans of human heads. Of those, we used 397 to generate training images, 40 for validation and 40 for testing of the neural network. The volumes have a resolution of $512\times512$ per slice and have between 101 and 645 slices per volume. 

From the volumes above, we generate 100K training images from random viewpoints and with randomized transfer functions (TF). The viewpoints are drawn at random from a uniform distribution of viewing directions with a random distance between $2.5$ and $3.0$ to the volume center. This configuration  places the volume around the center of the view frustum. The scenes are rendered with a fixed far plane distance $f=5.0$, and we define our depth $d$ as the normalized distance to the camera $c$ for a sample position $p$: \hspace{1mm} $d = |p  - c|_2^2 / f$

Note that with the proposed strategy we do not generate a very large variety in absolute depth values, as most images show visible structure roughly in the range $d \in [0.3, 0.7]$, however as noted before, the actual absolute values are not as relevant for our application, as they are not related to real-world units and can be re-scaled to match the desired scene extents.

In order to vary the appearance, and to control the amount of opacity in the renderings, we propose a transfer function generation scheme, similar to the approach proposed by Engel\etal~\cite{engel2020dvao}. This scheme generates piece-wise linear 1D transfer functions using randomly generated trapezoids, representing "peaks". For our datasets, we generate TFs containing between one and three of such peaks, while randomizing their widths (along intensity space) and heights (assigned opacity) to achieve a high variation of renderings with a controllable \emph{amount of opacity} in the rendered images. Controlling the amount of opacity is relevant to investigate how the amount of semi-transparent structures complicates the monocular depth estimation task in comparison to fully opaque scenes. Hence, we generate two different datasets:

\vspace{2mm}
\noindent\textsc{Opaque}: \hfill  $<$ 10\% of pixels influenced by \emph{semi-transparency} \\
\textsc{Transparent}: \hfill $>$ 20\% of pixels influenced by \emph{semi-transparency} \\
\vspace{-1mm}

Hereby, we classify a pixel as influenced by \emph{semi-transparency} if its accumulated opacity is in the interval $[0.02, 0.9]$, where an opacity of $1.0$ corresponds to fully opaque and $0.0$ to fully transparent. For the generation of the \textsc{Transparent} dataset, we decreased the maximum height of the generated trapezoid peaks compared to \textsc{Opaque} from $0.6$ to $0.4$. During generation, we reject all samples that do not satisfy the \emph{amount of opacity} condition. The resulting \textsc{Opaque} dataset has an average of $5.2\%$ \emph{semi-transparent} pixels and \textsc{Transparent} has an average of $29.8\%$ \emph{semi-transparent} pixels.
During training we further apply the standard image augmentation techniques employed by the respective depth-estimation method.

\begin{figure*}[tb]
    \centering
    \begin{subfigure}{0.2435\textwidth}  
        \includegraphics[width=\textwidth, height=5.33cm]{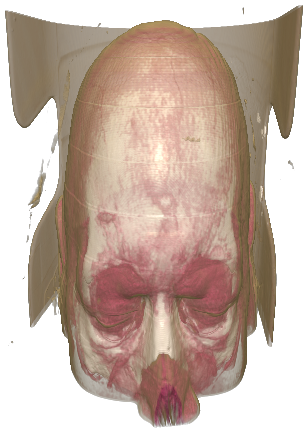}
        \caption{Input Color Image}
    \end{subfigure}%
    \hfill
    \begin{subfigure}{0.72\textwidth}
    \begin{subfigure}{0.19\textwidth}
        \includegraphics[width=\textwidth, height=3cm]{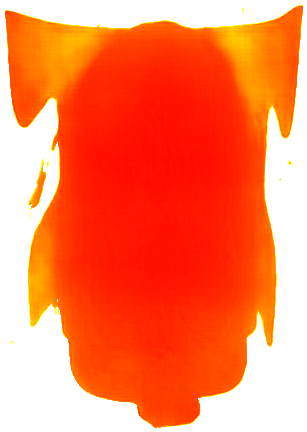}
    \end{subfigure}%
    \hfill
    \begin{subfigure}{0.19\textwidth}
        \includegraphics[width=\textwidth, height=3cm]{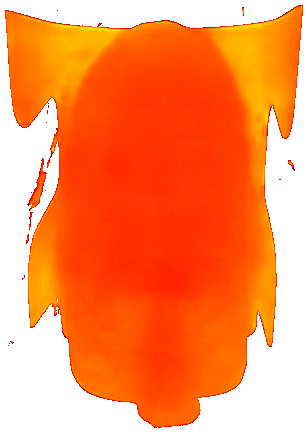}
    \end{subfigure}%
    \hfill
    \begin{subfigure}{0.19\textwidth}
        \includegraphics[width=\textwidth, height=3cm]{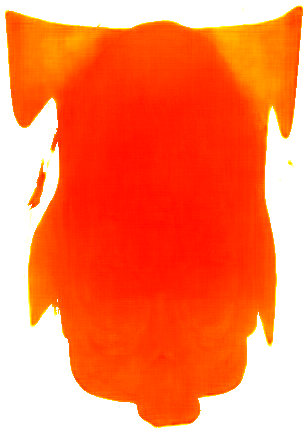}
    \end{subfigure}%
    \hfill
    \begin{subfigure}{0.19\textwidth}
        \includegraphics[width=\textwidth, height=3cm]{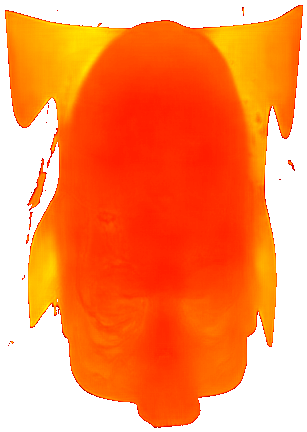}
    \end{subfigure}%
    \hfill
    \begin{subfigure}{0.19\textwidth}
        \includegraphics[width=\textwidth, height=3cm]{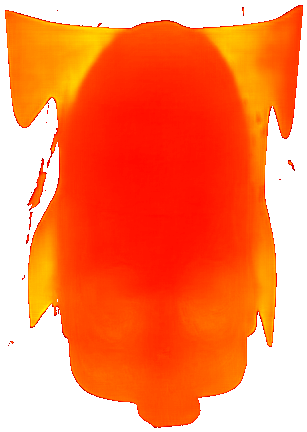}
    \end{subfigure}%
    \\
    \begin{subfigure}{0.19\textwidth}
        \includegraphics[width=\textwidth, height=3cm]{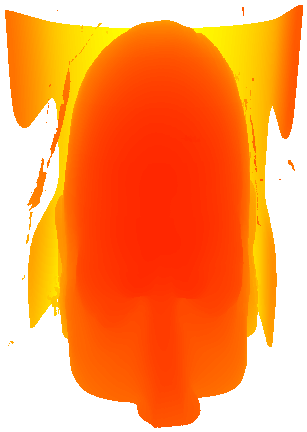}
        \caption{FirstHit}
    \end{subfigure}%
    \hfill
    \begin{subfigure}{0.19\textwidth}
        \includegraphics[width=\textwidth, height=3cm]{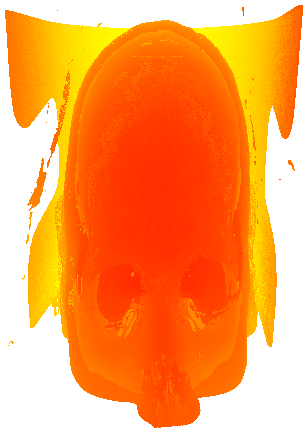}
        \caption{MaxOpacity}
    \end{subfigure}%
    \hfill
    \begin{subfigure}{0.19\textwidth}
        \includegraphics[width=\textwidth, height=3cm]{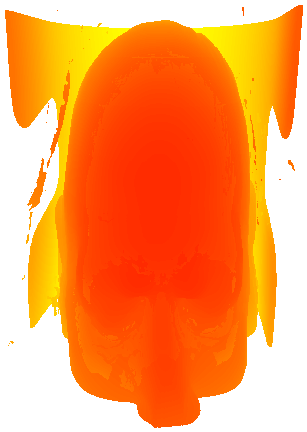}
        \caption{MaxGradient}
    \end{subfigure}%
    \hfill
    \begin{subfigure}{0.19\textwidth}
        \includegraphics[width=\textwidth, height=3cm]{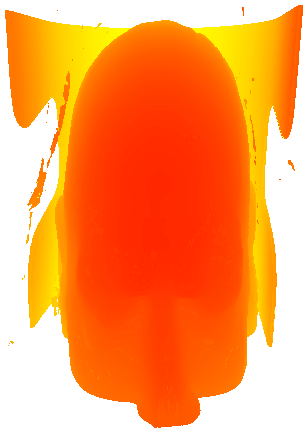}
        \caption{WYSIWYP}
    \end{subfigure}%
    \hfill
    \begin{subfigure}{0.19\textwidth}
        \includegraphics[width=\textwidth, height=3cm]{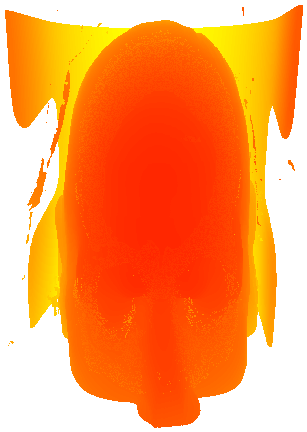}
        \caption{MIDA}
    \end{subfigure}%
    \end{subfigure}%
    \caption{\textbf{Qualitative monocular depth estimation results} on semi-transparent rendering (\emph{a}) for different \emph{depth techniques} (\emph{b-f}). Prediction of the \textsc{BTS} for the respective depth technique (\emph{top row}), and ground truth (\emph{bottom row}).}
    \label{fig:monodepth-results}
    \vspace{-0.2cm}
\end{figure*}

\subsection{Depth Techniques}\label{sec:depth-techniques}
Due to the ambiguity of surfaces in semi-transparent renderings, it is notoriously difficult to find a definitive depth value for a given pixel in such renderings. Prior work on volume picking offers a variety of approaches that are designed to find "relevant" surfaces or objects from a human perspective~\cite{wiebel2011perception}. Together with some common choices, we compare a total of five \emph{depth techniques} that we include in our experiments. Assuming a ray $r(\bm{t}) = c + \bm{t}\cdot\vec{r}$ with ray direction $\vec{r}$, distance along the ray $\bm{t}$ and local and accumulated opacities $\bm\sigma_r(\bm{t})$ and $\bm\sigma^*_r(\bm{t})$, we use the \emph{depth techniques} below (compare Fig.~\ref{fig:depth-technique-comparison}). If multiple depth values qualify for a technique, we use the smallest.

\noindent{\textbf{FirstHit.}} 
This is the simplest strategy. The depth of a pixel is given by the position of the first sample that has a non-zero opacity along a view ray. This strategy completely discards semi-transparency and essentially treats the scene like an opaque object. 
$$ \bm{d}_\text{FH} = \min_{\bm{t}} \left\{ \bm{t} \hspace{.5mm} | \hspace{.5mm} \bm{\sigma}_r(\bm{t}) > 0 \right\}$$

\noindent{\textbf{MaxOpacity.}} 
This strategy corresponds to taking the depth of the sample with the maximum opacity along a ray, i.e., the depth of the sample used for \emph{Maximum Intensity Projection} (MIP).
$$ \bm{d}_\text{MO} = \argmax_{\bm{t}} \bm{\sigma}_r(\bm{t})$$

\noindent{\textbf{MaxGradient.}}
The MaxGradient strategy looks for the steepest change in accumulated opacity along a ray. Similar to the following methods, this favors early samples along the rays.
$$ \bm{d}_\text{MG} = \argmax_{\bm{t}} \frac{\partial \bm{\sigma}^*_r(\bm{t})}{\partial \bm{t}}$$

\noindent{\textbf{WYSIWYP.}}
This approach follows the picking strategy presented by Wiebel\etal~\cite{wiebel2012wysiwyp}. This strategy finds \emph{intervals} along each ray, defined by the zero-crossing of the 2nd order derivative of the accumulated opacity. An interval begins when the 2nd derivative becomes positive after being negative or zero. The interval ends after the 2nd derivative has plateaued at $0$ again, corresponding to regions without opacity accumulation. After defining intervals, the chosen depth value is the distance to the start of the interval that increased the accumulated opacity the most.
\begin{align*}
\bm{d}_\text{WYSIWYP} &= \argmax_{\bm{t}_\text{s,i}} \left( \bm\sigma^*_r(\bm{t}_\text{e,i}) - \bm\sigma^*_r(\bm{t}_\text{s,i}) \right) \\
\bm{t}_\text{s,i} &= \min_{\bm{t} > \bm{t}_\text{e, i-1}} \left\{  \bm{t} \hspace{1mm} \big| \hspace{1mm} \frac{\partial^2 \bm\sigma^*_r(\bm{t})}{\partial \bm{t}^2} = 0 ,\hspace{1mm} \frac{\partial^3 \bm\sigma^*_r(\bm{t})}{\partial \bm{t}^3} > 0  \right\} \\
\bm{t}_\text{e,i} &= \hspace{1mm} \min_{\bm{t} > \bm{t}_\text{s,i}} \hspace{1mm} \left\{  \bm{t} \hspace{1mm} \big| \hspace{1mm} \frac{\partial^2 \bm\sigma^*_r(\bm{t})}{\partial \bm{t}^2} = 0 ,\hspace{1mm} \frac{\partial^3 \bm\sigma^*_r(\bm{t})}{\partial \bm{t}^3} = 0  \right\}
\end{align*}
Here $\bm{t}_s, \bm{t}_e$ denote the \emph{start} and \emph{end} depth of the interval $i$, and the first interval begins at $\bm{t}_\text{s,i} = \bm{d}_\text{FH}$. The intervals are illustrated on top of Fig.~\ref{fig:depth-technique-comparison}.

\noindent{\textbf{MIDA.}}
Lastly, we took inspiration from Maximum Intensity Difference Accumulation proposed by Bruckner\etal~\cite{bruckner2009mida}. MIDA is a compositing technique to combine benefits of DVR and MIP. In their work, they define a $\beta$ parameter that trades some already accumulated opacity along a ray for new, highly relevant (in a MIP sense) structures. To find a depth value, we use the sample with the lowest $\beta$ parameter, i.e., the sample that increases over the previous MIP (up to this sample) the most. 
\begin{align*}
    \bm{d}_\text{MIDA} &= \argmax_{\bm{t}} \left( \bm\sigma_r(\bm{t}) - \max_{\bm{\tau} < \bm{t}} \bm\sigma_r(\bm{\tau}) \right)
\end{align*}

\subsection{Monocular Depth Estimation}
As a first step, we investigate whether neural nets can predict depth under the above requirements. For this, we compare five SotA approaches that have shown great success on real-world data. 

In the following experiments, we use the method of Laina~\etal\cite{laina2016deeper} (\textsc{LAINA}), one of the first deep convolutional neuronal networks for monocular depth estimation.  Further, we use the well-known deep ordinal regression network (\textsc{DORN}), the first approach formulating depth estimation as classification problem. Additionally, we picked the approaches of Lee~\etal\cite{lee2019bts}, a multiscale approach (\textsc{BTS}) and Yin~\etal\cite{yin2019vnl} using virtual normals to enforce geometric constraints (\textsc{VNL}). Both approaches show strong performance on common depth estimation datasets. Finally, we adopt the network of Ranftl~\etal\cite{ranftl2020midas}(\textsc{MIDAS}), which yields robust depth predictions for a combination of multiple datasets, while also being scale and shift invariant.
In this task, we train the neural nets to predict a depth map for a semi-transparent rendering from just an RGB color image. 

To adapt this task to semi-transparent images, we have to make minor adjustments to the originally proposed neural nets. One of the adjustments is to compensate for the fact that the synthetically generated renderings typically do not encode real-world sizes, but rather encode \emph{depth} as a normalized distance within an enclosed view frustum. Specifically, we adapt the output activation functions, which are usually rectified linear units (ReLU), to sigmoid activations. This scales the network's output to $(0, 1)$, aligning it with the normalized depth we require. We then trained the networks with their proposed training procedure, including data augmentation, and losses on both the \textsc{Opaque} and \textsc{Transparent} dataset. We report our results in Sec.~\ref{sec:exp-monodepth}.

\begin{table*}[!htb]
    \hspace{-.02\textwidth}
    \resizebox{1.04\textwidth}{!}{%
    \begin{tabular}{l ccccc ccccc ccccc ccccc}
    \toprule 
    \textsc{\textbf{Depth}}  & 
    \multicolumn{4}{c}{\textsc{FirstHit}} & 
    \multicolumn{4}{c}{\textsc{MaxOpacity}} & 
    \multicolumn{4}{c}{\textsc{MaxGradient}} & 
    \multicolumn{4}{c}{\textsc{WYSIWYP}} &
    \multicolumn{4}{c}{\textsc{MIDA}} \\
    \cmidrule(lr){2-5}
    \cmidrule(lr){6-9}
    \cmidrule(lr){10-13}
    \cmidrule(lr){14-17}
    \cmidrule(lr){18-21}
    & $\bm\delta_1\uparrow$ & $\bm\delta_2\uparrow$ & $\bm\delta_3\uparrow$ & $\bm{L}_1\downarrow$  
    & $\bm\delta_1\uparrow$ & $\bm\delta_2\uparrow$ & $\bm\delta_3\uparrow$ & $\bm{L}_1\downarrow$   
    & $\bm\delta_1\uparrow$ & $\bm\delta_2\uparrow$ & $\bm\delta_3\uparrow$ & $\bm{L}_1\downarrow$ 
    & $\bm\delta_1\uparrow$ & $\bm\delta_2\uparrow$ & $\bm\delta_3\uparrow$ & $\bm{L}_1\downarrow$
    & $\bm\delta_1\uparrow$ & $\bm\delta_2\uparrow$ & $\bm\delta_3\uparrow$ & $\bm{L}_1\downarrow$   \\
    \rule{0pt}{1ex} &&&&& &&&&& &&&&& &&&&& \\
    \multicolumn{2}{l}{\textsc{\textbf{Opaque}}} &&&& &&&&& &&&&& &&&&& \\
    \midrule 
    \textsc{BTS}    &  .963    &  .981   &  .987   &  .026      
           &  .937    &  .979   &  .986   &  .038      
           &  \textbf{.961}    &  .981   &  .987   &  .029      
           &  \textbf{.958}    &  .981   &  .986   &  \textbf{.029}      
           &  \textbf{.952}    &  .980   &  .986   &  .031   \\ 
           
    \textsc{VNL}    &  \textbf{.966}    &  \textbf{.994}   &  \textbf{.998}   &  .033       
           &   .935   &  \textbf{.983}   &  \textbf{.997}   &  .043      
           &   .950   &  \textbf{.995}   &  \textbf{.999}   &  .048      
           &   .954   &  \textbf{.993}   &  \textbf{.999}   &  .046      
           &   .872   &  \textbf{.986}   &  \textbf{.997}   &  .064   \\ 
           
    \textsc{LAINA}  &  .927    &  .958   &  .970   &  \textbf{.021}    
           &  .904    &  .954   &  .968   &  .030       
           &  .925    &  .958   &  .970   &  .022      
           &  .923    &  .957   &  .970   &  .023       
           &  .923    &  .958   &  .970   &  .023    \\ 
           
    \textsc{DORN}   &  .859    &  .948   &  .960   &  .050       
           &  .578    &  .877   &  .957   &  .091       
           &  .626    &  .891   &  .955   &  .084       
           &  .853    &  .946   &  .958   &  .051       
           &  .839    &  .946   &  .958   &  .054    \\ 
           
    \textsc{MIDAS}  &  .955    &  .977   &  .984   &  .022      
           &  \textbf{.942}    &  .976   &  .985   &  \textbf{.026}      
           &  .959    &  .979   &  .985   &  \textbf{.021}      
           &  .947    &  .976   &  .983   &  .030      
           &  .950    &  .980   &  .986   &  \textbf{.021}   \\ 
    \rule{0pt}{1ex} &&&&& &&&&& &&&&& &&&&& \\
    \multicolumn{2}{l}{\textsc{\textbf{Transparent}}} &&&& &&&&& &&&&& &&&&& \\
    \midrule 
    \textsc{BTS}    &   \textbf{.964}   &  \textbf{.985}   &   .989  &  .032     
           &   .883   &  \textbf{.977}   &   .988  &  \textbf{.052}     
           &   \textbf{.954}   &  \textbf{.984}   &   .989  &  .034       
           &   \textbf{.925}   &  .982   &  .988    &  .043     
           &   \textbf{.936}   &  \textbf{.982}   &  .989    &  .038  \\ 
               
    \textsc{VNL}    &   .947   &  .982   &   \textbf{.990}  &  .039     
           &   .833   &  .946   &   \textbf{.990}  &  .063      
           &   .655   &  .977   &   \textbf{.996}  &  .079     
           &   .891   &  \textbf{.983}   &   \textbf{.997}  &  .056     
           &   .845   &  .975   &   \textbf{.994}  &  .065  \\ 
           
    \textsc{LAINA}  &   .934   &  .966   &   .977  &  \textbf{.022}     
           &   .865   &  .958   &   .976  &  .046      
           &   .926   &  .965   &   .977  &  \textbf{.024}      
           &   .909   &  .963   &   .977  &  \textbf{.032}     
           &   .918   &  .966   &   .977  &  \textbf{.030}  \\ 
           
    \textsc{DORN}   &   .869   &  .959   &   .970  &  .047     
           &   .641   &  .927   &   .968  &  .083     
           &   .890   &  .959   &   .969  &  .047     
           &   .795   &  .953   &   .967  &  .063     
           &   .757   &  .946   &   .970  &  .065  \\ 
           
    \textsc{MIDAS}  &  .949    &  .980   &  .987   &  .029     
           &  \textbf{.904}    &  \textbf{.977}   &  .989   &  .053     
           &  .935    &  .979   &  .987   &  .034     
           &  .923    &  .980   &  .988   &  .038     
           &  .931    &  .979   &  .989   &  .036  \\ 
    \bottomrule 
    \end{tabular}
    }
    \caption{\textbf{Quantitative monocular depth estimation results} for the different \emph{depth techniques}, datasets and depth estimation networks (\textbf{rows}). We show depth predictions for the best method on the \textsc{Transparent} dataset in Fig.~\ref{fig:monodepth-results}. }
    \label{tab:monodepth}
\end{table*}

\subsection{Monocular Layered Representation Prediction}\label{sec:rgbad}
Since we found that the standard monocular depth estimation task works quite well on semi-transparent data, we further extended the approach, to predict additional information on top of the depth map. Given, that we can predict depths from relevant surfaces in the semi-transparent renderings, we now further ask the network to separate the color and opacity of the scene at those depths. That means we let the network predict the accumulated color and opacity in \emph{front} of the predicted depth, as well as in the \emph{back} of the predicted depth, corresponding to a 2-layer representation. In the following, we discuss our choice of the depth technique and depth estimator for this task and how they need to be adapted. Lastly, we present three novel ideas to further improve our results.

\noindent{\textbf{Choice of depth technique.}}
For these experiments, we chose to use only the \textsc{BTS}~\cite{lee2019bts} network in combination with the WYSIWYP depth. The reason is that depth techniques like \emph{MaxOpacity} and \emph{MaxGradient} result in very sparse \emph{front} layers after separation, while using \emph{FirstHit} would be meaningless, as \emph{front} would be $0$ by definition. MIDA and WYSIWYP both provide reasonable separation, however they both still coincide with \emph{FirstHit} depths for many of the pixels. In order to include those thin semi-transparent layers in the \emph{front}, we adjust the depths to be just behind that structure. While MIDA does not consider the thickness of structures, WYSIWYP readily provides full intervals. To include the first-hit structures, we modify the WYSIWYP approach to use the depth at the interval end, instead of the start, if the first interval would be chosen to represent the depth (compare Fig.~\ref{fig:depth-technique-comparison}).

\noindent{\textbf{Choice of neural net.}}
We chose \textsc{BTS}~\cite{lee2019bts} for our continued experiments for multiple reasons. Firstly, \textsc{VNL}~\cite{yin2019vnl} and \textsc{DORN}~\cite{fu2018dorn} disqualify for our further adaptations, because they are unable to scale computationally to the amount of output channels necessary. They both actually frame the depth estimation as a classification problem, mapping the classes to a discrete set of possible depths. This is very memory intensive and forbids a drastic increase in output size, as we propose it. From the remaining networks, \textsc{BTS} performs the best in terms of $\bm\delta_1$ on the \textsc{Transparent} dataset for WYSIWYP, which we consider the most meaningful metric for the task. We can also confirm empirically that \textsc{BTS} produces the best results and its training procedure is quite reliable.

\noindent{\textbf{Layered representation.}}  
We now ask the networks to separate the semi-transparent structures in a \emph{front} layer, before the predicted depth values, and a \emph{back} layer, behind the predicted depth, as illustrated in Fig.~\ref{fig:teaser}.
The \emph{front} layer can easily be retrieved during ray casting, by exporting the accumulated RGBA buffer when reaching the depth value to separate at. The \emph{back} can then be computed from the \emph{front} and the fully composited rendering $\bm{c}_\text{full}$ by inversion of the compositing step:
\begin{align*}
    \bm\alpha_\text{back}  = \frac{\bm\alpha_\text{full} - \bm\alpha_\text{front}}{1 - \bm\alpha_\text{front}}  , \hspace{5mm}
    \bm{c}_\text{back} = \frac{\bm{c}_\text{full} - \bm{c}_\text{front}}{1 - \bm\alpha_\text{front}} \cdot \frac{1}{\bm\alpha_\text{back}}
\end{align*}
The resulting prediction is essentially a layered representation of the rendered scene using two layers, that composites to the original input image. This layered representation allows for visualization-centric modifications of the input image, like compositing additional objects into the scene, as also demonstrated in Fig.~\ref{fig:teaser}, but should generally enable all types of modifications presented in prior works on these layered representations~\cite{shade1998layered,frey2013explorable,zellmann2012image,tikhonova2010explorable}.

In order to predict such layered representation, we modify the last layers of \textsc{BTS} to output a 10-channel image, instead of a single-channel image. The 10 channels consist of: 
\begin{itemize}
    \item The WYSIWYP depth map
    \item RGBA of the structures in \emph{front} of the depth map
    \item RGBA of the structures behind the depth map (\emph{back})
    \item The FirstHit depth map (to compute the thickness of \emph{front})
\end{itemize}
With the inclusion of color layers in the prediction, we also remove any color-related data augmentation that is part of the depth estimator's training procedure.
We refer to the resulting network as \emph{baseline} (compare Fig.~\ref{fig:rgbad-ablation}). While this baseline suffers from blurry layers and often fails to actually compose to the original image, we introduce the following three learning extensions.

\noindent{\textbf{Residual image output.}}
In the baseline, we find that predicted \emph{front} and \emph{back} layers suffer from blurriness, which we attribute to the decoder of the network. This decoder part is prone to introducing blur, and the skip connections from early layers now have to transport a lot more information compared to when just predicting depth. As a solution to the blurriness, we propose \emph{Residual Image Outputs}. Instead of directly predicting the \emph{front} and \emph{back} RGBA layers, we predict the residual to the input image. Using a $\tanh$-activation we predict how much color and alpha must be removed from or added to the input image, while preserving the input pixel for weakly activated neurons. For the alpha channels, we predict the residuals to the mean of the input color channels, as the input image's alpha is unavailable during inference. The idea is described by the following equation, given neural net $\bm{F}$, parameterized by $\Theta$ and input image $x$:

$$ \bm{\hat c}_\text{old} =  \bm{F}_\Theta (x),  \hspace{5mm} \bm{\hat c}_\text{new} = \bm{F}_\Theta (x) + x $$

\noindent{\textbf{Compositing loss.}}
In addition, we introduce a novel, visualization-centered loss function, which we call \emph{Compositing Loss}. The idea of the loss is to enforce consistency in the layered image representation, by enforcing the \emph{front} and \emph{back} layers to alpha-composite to the original input rendering, as described by the following equation, given color and opacity predictions $\bm{\hat c}$ and $\bm{\hat \alpha}$, number of \emph{valid} pixels $N$ and ground truth color and opacity $\bm{c}$, $\bm{\alpha}$:

\begin{align*}
    \bm{\hat c}      &= \bm{c}_\text{front} \hspace{0.75mm} + (1 - \bm\alpha_\text{front}) \cdot \bm\alpha_\text{back} \cdot  \bm{c}_\text{back}\\
    \bm{\hat \alpha} &= \bm\alpha_\text{front} + (1 - \bm\alpha_\text{front}) \cdot  \bm\alpha_\text{back} \\
    \bm{L}_\text{Comp} &= \frac{1}{N} \sum_i^N |\bm{\hat c}_i - \bm{c}_i|_2^2 + |\bm{\hat \alpha}_i - \bm{\alpha}_i|_2^2
\end{align*}
Hereby we consider pixels as \emph{valid} if they have an $\bm\alpha > 0$, discarding the pixels showing only background, as they would dominate the loss due to the sparsity of our data. Note that this mask would be unavailable during inference, as we cannot expect to have an alpha channel for the input image. During inference we can instead mask the predicted output using the input color channels.

\noindent{\textbf{Front-Back-Divergence loss.}}
The last loss term $\bm{L}_\text{FBDiv}$ acts as a regularization to the model. Through the compositing loss, the model is encouraged to produce layers that composite to the original input image. Using this loss alone often results in predictions for the \emph{front} layer to be very similar to the \emph{back}, with just decreased opacity. As a consequence, the \emph{front} layer would often contain color of the structures that should be in \emph{back} instead.
To counter this effect we added the front-back-divergence loss term, which encourages the network to put structures of different color into different layers, by penalizing pixels with the same color in both \emph{front} and \emph{back}.
This front-back-divergence is implemented by minimizing cosine similarity of \emph{front} and \emph{back}:
$$ \bm{L}_\text{FBDiv} = \frac{1}{N} \sum_i^N \left< \bm{\hat c}_\text{i, front} , \bm{\hat c}_{i, back} \right>$$

\noindent{\textbf{Overall loss.}} The overall loss function to train the network for joint depth estimation \emph{and} color/alpha separation is a weighted sum of the \emph{SILog} loss~\cite{eigen2014depth} on the depth map only, a \emph{mean absolute error} ($\bm{L}_1$) and structured dissimilarity ($\bm{L}_\text{SSIM}$) on the color and alpha channels, and the compositing loss on the re-composited image, which backpropagates through the color and alpha predictions as well. Lastly the Front-Back-Divergence loss ($\bm{L}_\text{FBDiv}$) is computed on the color components of the \emph{front} and \emph{back} predictions.
 \begin{align*}
    \bm{L}_\text{SILog} &= \frac{1}{N} \sum_i^N (\log{\bm{d}_i} - \log{\bm{\hat d}_i})^2    -    
        \frac{1}{N^2} \left( \sum_i^N \log{\bm{d}_i} - \log{\bm{\hat d}_i} \right)^2 \\
    \bm{L}_\text{1, RGBA} &= \frac{1}{N} \sum_i^N | \bm{\hat c}_i - \bm{c}_i | + | \bm{\hat \alpha}_i - \bm{\alpha}_i | \\
    \bm{L}_\text{SSIM} &= 1 - \text{SSIM}(\bm{\hat c}, \bm{c}) \\
    \bm{L}_\text{total} &= \bm{L}_\text{1, RGBA} + \lambda_\text{Depth} \cdot \bm{L}_\text{SILog} +  \lambda_\text{Comp} \cdot \bm{L}_\text{Comp}  \\
    &\hspace{4mm} + \lambda_\text{FBDiv} \cdot \bm{L}_\text{FBDiv} + \lambda_\text{SSIM} \cdot \bm{L}_\text{SSIM}
 \end{align*}

Balancing the weights of this loss is important to achieve good results, especially because the individual loss terms have different value ranges. It seems most important to let the $\bm{L}_\text{SILog}$ dominate the loss magnitude, as the quality of the depth prediction empirically has the biggest influence on the end results, qualitatively. Furthermore, the $\bm{L}_\text{FBDiv}$ component will force the network to produce two very distinct colors in the \emph{front} and \emph{back} layers, resulting in layers mostly containing either red, green or blue, regardless of the color of the structures, making $\lambda_\text{FBDiv}$ another sensitive parameter. In our experiments we chose $\lambda_\text{Depth} = 10.0$, $\lambda_\text{Comp} = 2.0$, $\lambda_\text{SSIM} = 2.0$ and $\lambda_\text{FBDiv} = 0.2$ empirically based on the overall magnitude each of the losses has. With this setting the contribution of $\bm{L}_\text{SILog}$ to $\bm{L}_\text{total}$ is roughly the same as the sum of all losses based on color predictions.

\begin{figure}[htb]
    \centering
    \begin{subfigure}{0.25\linewidth}
        \includegraphics[width=\textwidth]{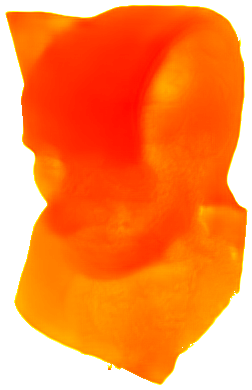}
        \caption{Prediction}
    \end{subfigure}%
    \begin{subfigure}{0.25\linewidth}
        \includegraphics[width=\textwidth]{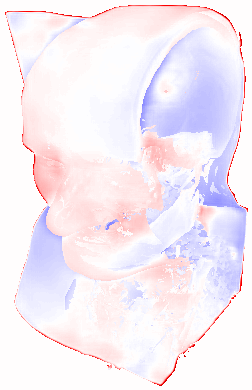}
        \caption{Depth Error}
    \end{subfigure}%
    \begin{subfigure}{0.25\linewidth}
        \includegraphics[width=\textwidth]{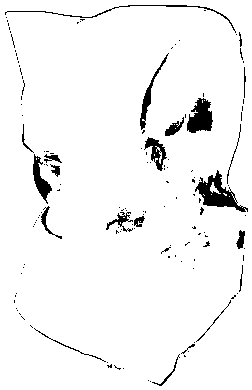}
        \caption{$\bm\delta_1$ Mask}
    \end{subfigure}%
    \begin{subfigure}{0.25\linewidth}
        \includegraphics[width=\textwidth]{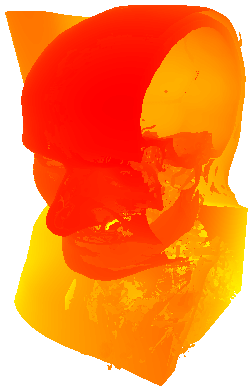}
        \caption{Ground Truth}
    \end{subfigure}%
    \\
    \caption{\textbf{Depth prediction errors}. Here (\emph{b}) shows absolute errors in the predicted depth. (\emph{c}) indicates the pixels' contributions to the $\bm\delta_1$ scores, showing pixels with $>25\%$ relative error in black. Most of the errors occur on either the object boundaries or on the boundary between different structures.}
    \label{fig:depth-error-images}
    \vspace{-3mm}
\end{figure}

\vspace{-2mm}
\section{Evaluation of Depth Estimators}\label{sec:experiments}
We have performed several experiments to investigate the monocular depth estimation performance, as well as the monocular layered representation predictions. The depth estimation experiments evaluate how well neural networks perform on images of semi-transparent renderings, specifically we
(1) measure the effect of the amount of transparency in the input images and (2) compare the different depth techniques presented in Sec.~\ref{sec:depth-techniques} to each other and how well they can be predicted by our networks.

Based on the results of those first experiments, we choose a neural net and depth technique for our second set of experiments, investigating the monocular layered representation prediction. In this second set of experiments, we compare the proposed baseline for the separation task with the addition of the proposed \emph{Residual Image Outputs}, \emph{Compositing Loss} and \emph{Front-Back-Divergence Loss}.

\noindent{\textbf{Metrics.}}
We measure the performance of the networks using metrics commonly used for monocular depth estimation, as well as image similarity metrics for the RGBA layers:
\vspace{-0.5mm}
\begin{itemize}
    \item $\bm\delta$-metric: The $\bm\delta_1, \bm\delta_2, \bm\delta_3$ metrics measure the percentage of \emph{valid} predictions ($\bm{\hat d_i}$) within a certain margin of relative error (here $25\%$) compared to the true depths ($\bm{d_i}$): 
    $$\bm\delta_k = \frac{1}{N} \left| \left\{ i \hspace{1mm} \big| \hspace{1mm} \max \left( \frac{\bm{d}_i}{\bm{\hat d}_i}, \frac{\bm{\hat d}_i}{\bm{d}_i}\right) < \bm\lambda^k \right\} \right| \text{ with } \bm\lambda=1.25$$
    \item $\bm{L}_1$-metric: \emph{mean absolute error} of \emph{valid} pixels
    \item SSIM~\cite{wang2004ssim}: The structured similarity index:
    $$
    \text{SSIM}(a,b) =  \frac{(2\mu_a\mu_{b} + \epsilon_1)(2\sigma_{ab} + \epsilon_2)}{(\mu_a^2 + \mu_{b}^2 + \epsilon_1)(\sigma_a^2 + \sigma_{b}^2 + \epsilon_2)} $$
    Hereby $\epsilon_i$ are small constants for numerical stability and $\mu_a$, $\sigma_a^2$ and $\sigma_{ab}$ are the means, variances and covariances within a local neighborhood for all pixels of $a$ (and $b$) respectively.
\end{itemize}
\vspace{-0.5mm}
For the $\bm\delta$-metrics and $\bm{L}_1$ we again only consider the $N$ \emph{valid} pixels, indexed by $i$, that are not part of the background.

\begin{table}[!ht]
  \centering
  \begin{tabular}{lccccc}
    \toprule
    \textsc{\textbf{Layer}} & \textsc{Metric} & \textsc{Base} & \textsc{+R} & \textsc{+R+C} & \textsc{+R+C+FB} \\
    \midrule
    Front & \multirow{3}{*}{SSIM $\uparrow$}
                &  .731 & .744 & .800 & \textbf{.802} \\
    Back       &&  .689 & .827 & .798 & \textbf{.833} \\
    Composite  &&  .830 & .894 & .958 & \textbf{.961} \\

    \rule{0pt}{1mm} &&&&& \\ 
    Front & \multirow{3}{*}{$\bm{L}_1\downarrow$}
                & \textbf{.070} & .073 & .073 & .073 \\
    Back       && .080 & .087 & .081 & \textbf{.079} \\
    Composite  && .043 & .046 & \textbf{.025} & \textbf{.025} \\

\rule{0pt}{1mm} &&&&& \\ 
    \textsc{\textbf{Depth}} && \multicolumn{4}{c}{} \\
    \midrule
    FirstHit & \multirow{2}{*}{$\bm\delta_1\uparrow$}
              & \textbf{.981} & .980 & \textbf{.981} & \textbf{.981} \\
    WYSIWYP  && .964 & .963 & \textbf{.966} & .964 \\

    \rule{0pt}{1mm} &&&&& \\ 
    FirstHit & \multirow{2}{*}{$\bm{L}_1\downarrow$}
              & .015 & .015 & \textbf{.014} & \textbf{.014} \\
    WYSIWYP  && .021 & .022 & \textbf{.019} & .021 \\
    \bottomrule
  \end{tabular}
  \caption{\textbf{Ablation study} of the \emph{Residual Image Output} (+R), \emph{Compositing Loss} (+C) and \emph{Front-Back-Divergence} (+FB). Visual results in Fig.~\ref{fig:rgbad-ablation}.}
  \label{tab:rgbad-ablation}
  \vspace{-3mm}
\end{table}

\vspace{-2mm}
\subsection{Monocular Depth Estimation}\label{sec:exp-monodepth}
\noindent{\textbf{Experiment.}} In this experiment, we evaluate the applicability of five SotA monocular depth estimation approaches on the different depth techniques introduced in Sec.~\ref{sec:depth-techniques}. The five approaches we test are \textsc{DORN}~\cite{fu2018dorn}, \textsc{MIDAS}~\cite{ranftl2020midas}, \textsc{LAINA}~\cite{laina2016deeper}, \textsc{VNL}~\cite{yin2019vnl} and \textsc{BTS}~\cite{lee2019bts}. We chose this set of networks because they are top performers on the monocular depth estimation leaderboards and have publicly available implementations. The goal of this experiment is to find out if such approaches can be applied to semi-transparent images and how the degree of opacity impacts the neural nets. 
The quantitative results of this experiment are presented in Table~\ref{tab:monodepth}. Additionally, Fig.~\ref{fig:monodepth-results} shows the depth predictions of the \textsc{BTS} depth estimator, for each of the depth techniques. Lastly, Fig.~\ref{fig:depth-error-images} shows an error image, as well as a mask highlighting which pixels are considered correct or false for the $\bm\delta_1$-metric.

\noindent{\textbf{Discussion.}} The first thing to note from the experiment is that all approaches perform reasonably well on the presented data. Especially for the \emph{FirstHit} depth technique, we get impressive $\bm\delta$-scores, while the networks perform quite similar on the \textsc{Opaque} and \textsc{Transparent} datasets. In contrast, the remaining depth techniques show a bigger drop in performance when decreasing the amount of opacity. The \emph{MaxOpacity} depth shows the most severe decrease in $\bm\delta$-scores, and we think this is the case, because it is the least biased towards the front and thus closer to the \emph{FirstHit} depth. \emph{MaxOpacity} is completely invariant to the position of structures along the ray, whereas all other techniques take the accumulated opacity into account, which increases faster at the beginning of a ray, because there has not occurred so much absorption yet.

\subsection{Monocular Layered Representation Prediction}\label{sec:exp-rgbad}
Based on the previous experiments and the adaptability reasons discussed in Sec.~\ref{sec:rgbad}, we choose \textsc{BTS} in combination with the WYSIWYP depth technique for the upcoming experiments. The fully extended BTS model runs in around $1s$ on an Intel i7-8700K (CPU) and around $0.1s$ on an RTX 2070 (GPU).

\begin{figure}
    \centering
    \begin{subfigure}{.2\linewidth}
        \includegraphics[width=\textwidth, height=1.8cm]{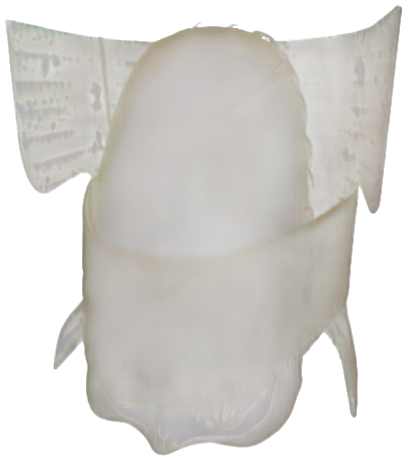}
        \\
        \includegraphics[width=\textwidth, height=1.8cm]{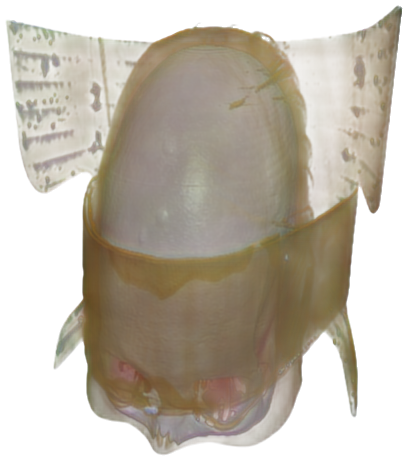}
        \\
        \includegraphics[width=\textwidth, height=1.8cm]{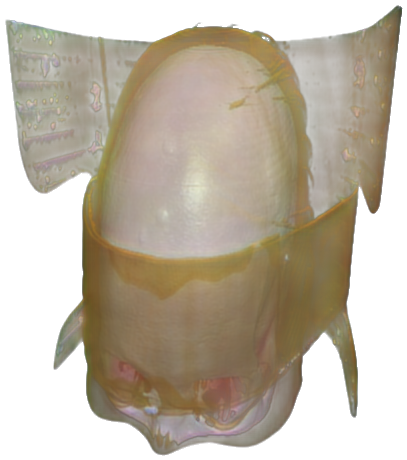}
        \caption{Baseline}
    \end{subfigure}%
    \begin{subfigure}{.2\linewidth}
        \includegraphics[width=\textwidth, height=1.8cm]{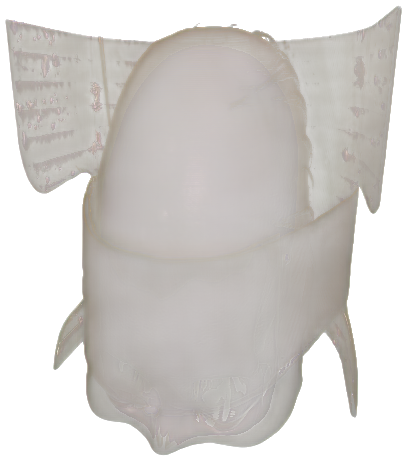}
        \\
        \includegraphics[width=\textwidth, height=1.8cm]{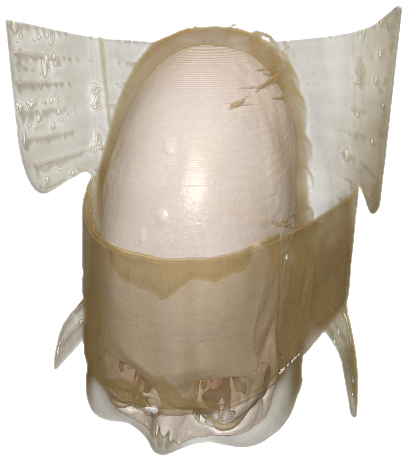}
        \\
        \includegraphics[width=\textwidth, height=1.8cm]{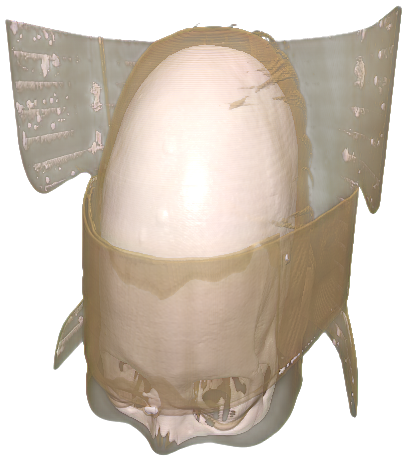}
        \caption{+R}
    \end{subfigure}%
    \begin{subfigure}{.2\linewidth}
        \includegraphics[width=\textwidth, height=1.8cm]{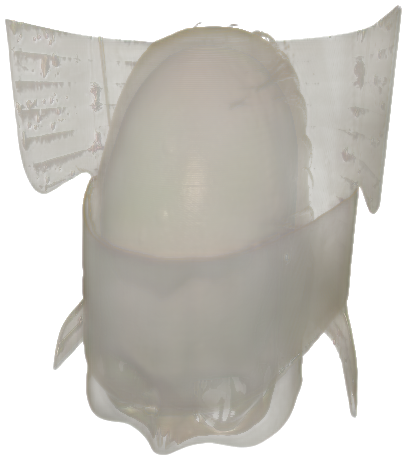}
        \\
        \includegraphics[width=\textwidth, height=1.8cm]{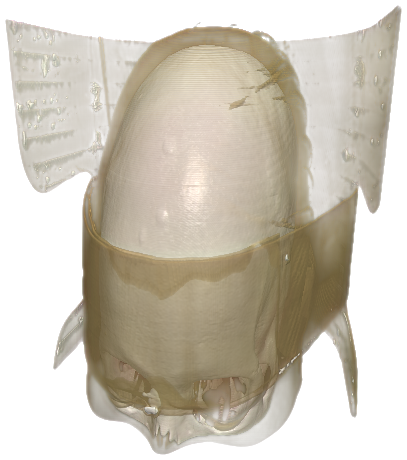}
        \\
        \includegraphics[width=\textwidth, height=1.8cm]{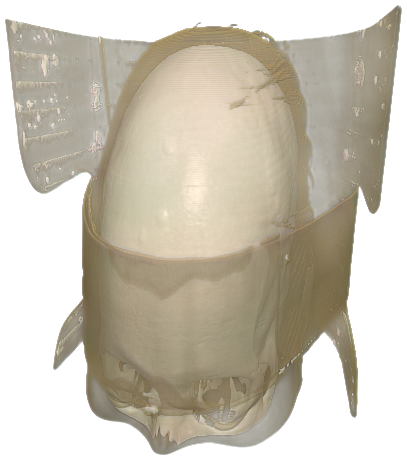}
        \caption{+R+C}
    \end{subfigure}%
    \begin{subfigure}{.2\linewidth}
        \includegraphics[width=\textwidth, height=1.8cm]{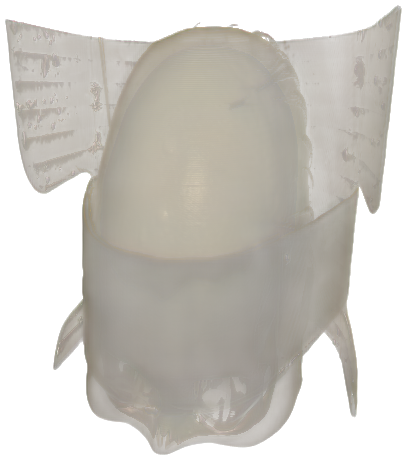}
        \\
        \includegraphics[width=\textwidth, height=1.8cm]{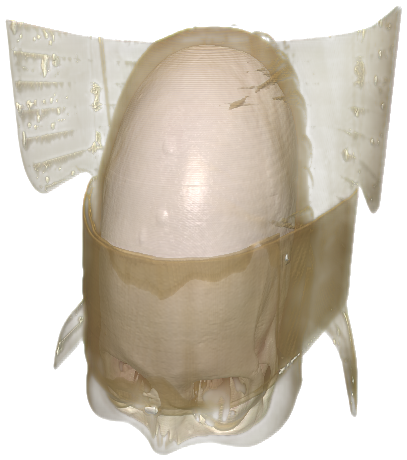}
        \\
        \includegraphics[width=\textwidth, height=1.8cm]{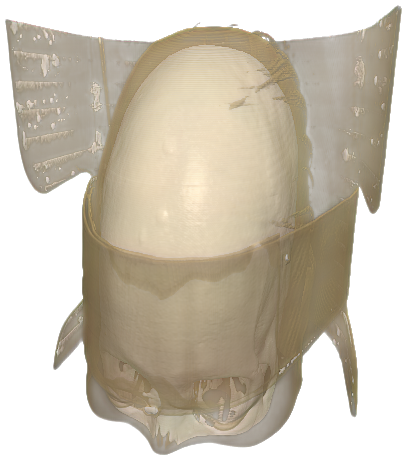}
        \caption{+R+C+FB}
    \end{subfigure}%
    \begin{subfigure}{.2\linewidth}
        \includegraphics[width=\textwidth, height=1.8cm]{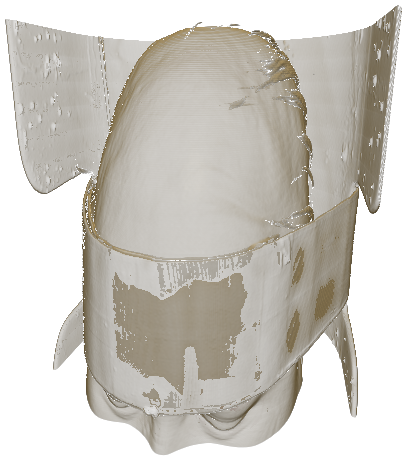}
        \\
        \includegraphics[width=\textwidth, height=1.8cm]{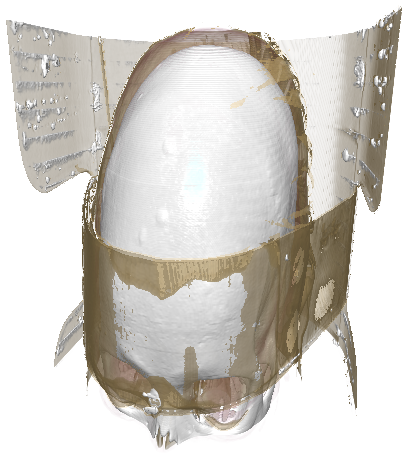}
        \\
        \includegraphics[width=\textwidth, height=1.8cm]{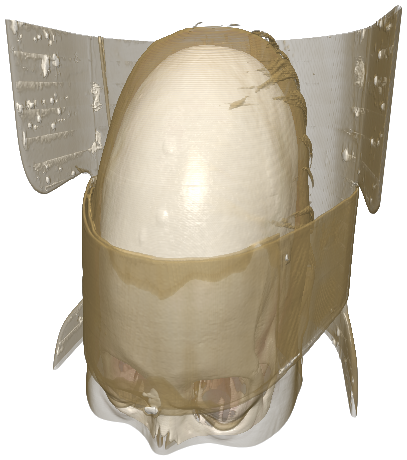}
        \caption{GT}
    \end{subfigure}%
    \caption{\textbf{Ablations for monocular layered representation prediction.} We compare the baseline neural net (\emph{a}) with additional Residual Image Output (+R), with additional Compositing Loss (+R+C), and with all three proposed enhancements (+R+C+FB), to the Ground Truth (\emph{e}). We show the Front layer (\emph{top}), the Back layer (\emph{middle}), and the Re-Composited Image (\emph{bottom}).}
    \label{fig:rgbad-ablation}
    \vspace{-1mm}
\end{figure}

\noindent{\textbf{Experiment.}}
This experiment evaluates our proposed extended version of \textsc{BTS} quantitatively. We predict the \emph{Front}, \emph{Back} and \emph{Depth} layers from an input RGB image (see Fig.~\ref{fig:teaser}) and compare them with the respective ground truths using SSIM and $\bm{L}_1$. The ablation study for our different variants, with \emph{Residual Image Outputs}, \emph{Compositing Loss} and \emph{Front-Back-Divergence} can be seen in Table~\ref{tab:rgbad-ablation} with the  layered representations illustrated in Fig.~\ref{fig:rgbad-ablation}.

\noindent{\textbf{Discussion.}}
The resulting layers shown in Fig.~\ref{fig:rgbad-ablation} make clear that our baseline approach, with just the number of output layers of the network adapted, suffers from blurry predictions. The proposed residual image outputs (\textbf{+R}) in contrast are much crisper, as the network receives a pixel-precise signal to its last layer, making it easier to keep sharp boundaries. This is also confirmed by the increase in SSIM of \textbf{+R} over the baseline. But still, the \textbf{+R} variant's re-composition is visually quite different from the ground truth. Adding the proposed compositing loss to the variant (\textbf{+R+C}) significantly improves the results in this regard. Lastly we found that with this setting the network would often predict a very similar \emph{Front} and \emph{Back} layer, that composites to the original image, but does not properly separate the structures within the color layers. Applying our proposed \emph{Front-Back-Divergence} loss (\textbf{+R+C+FB}) further improves the image quality, especially for the back layer, while maintaining the quality of the composition. Lastly note that all of our proposed ablations maintain a very similar depth prediction performance, while strongly improving the visual quality of the color layers.

\begin{figure}[!bt]
    \centering
    \begin{subfigure}{0.33\linewidth}
        \includegraphics[width=\textwidth]{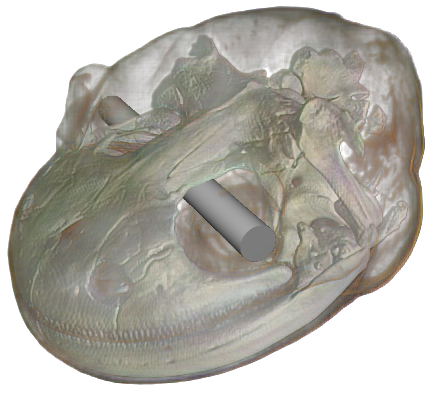}
        \caption{Re-Composited}
    \end{subfigure}%
    \begin{subfigure}{0.33\linewidth}
        \includegraphics[width=\textwidth]{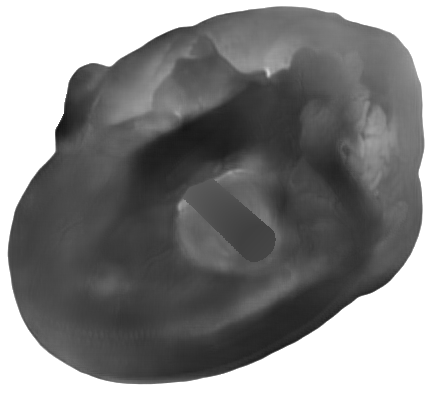}
        \caption{Depth}
    \end{subfigure}%
    \begin{subfigure}{0.33\linewidth}
        \includegraphics[width=\textwidth]{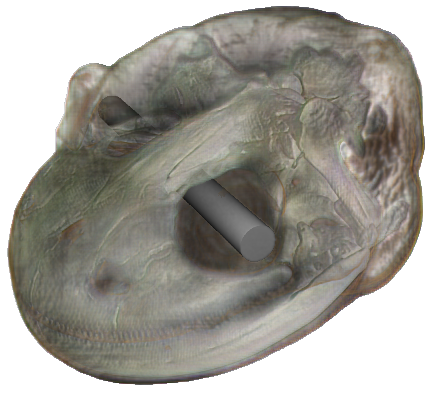}
        \caption{with AO}
    \end{subfigure}%
    \caption{\textbf{Out of distribution example.} Although our training set only contains human CT data, our method also generalizes to other specimen, like the salamander.}
    \label{fig:rgbad-comp-nonhead2}
    \vspace{-4mm}
\end{figure}

\begin{figure*}[!ht]
    \centering
    \begin{subfigure}{0.211\linewidth}
        \includegraphics[width=\textwidth]{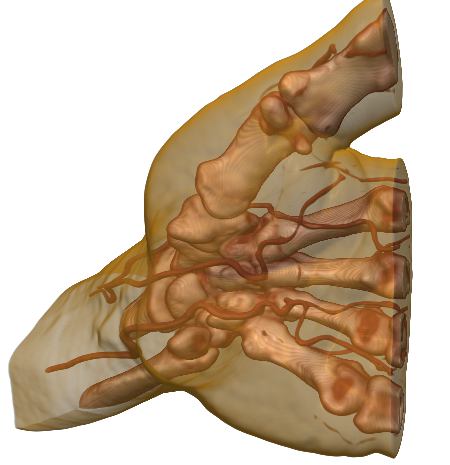}
        \caption{Input RGB}
    \end{subfigure}%
    \hfill
    \begin{subfigure}{0.1515\linewidth}
        \includegraphics[width=\textwidth]{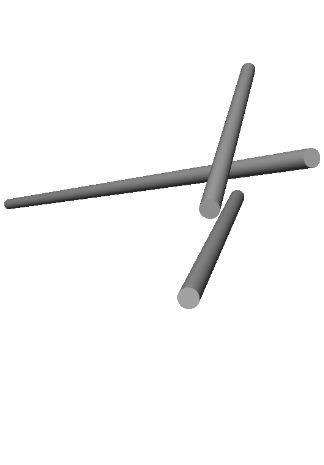}
        \caption{Cylinders only}
    \end{subfigure}%
    \begin{subfigure}{0.211\linewidth}
        \includegraphics[width=\textwidth]{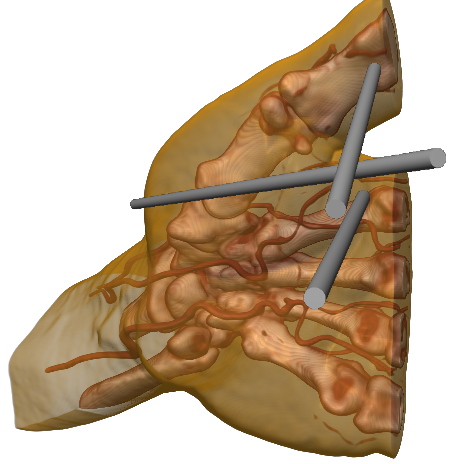}
        \caption{Naive Compositing}\label{fig:rgbad-comp-hand-naive}
    \end{subfigure}%
    \begin{subfigure}{0.211\linewidth}
        \includegraphics[width=\textwidth]{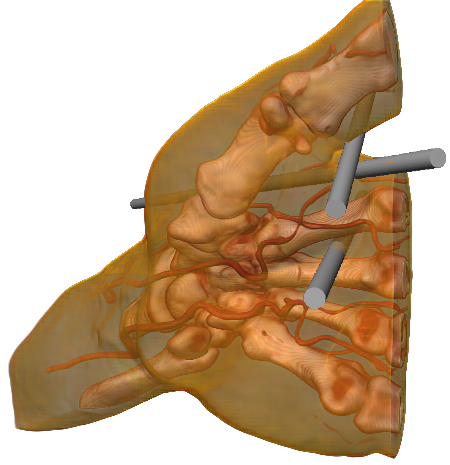}
        \caption{Ours}
    \end{subfigure}%
    \begin{subfigure}{0.211\linewidth}
        \includegraphics[width=\textwidth]{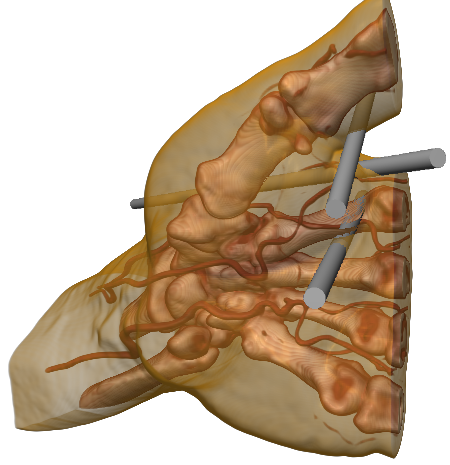}
        \caption{Ground Truth}
    \end{subfigure}%
    \caption{\textbf{Depth and alpha compositing}. Here we compose the input image with a set of cylinders. The naive approach just renders the cylinders on top of the hand, fully ignoring the geometry and transparency. Using our approach, we can correctly composite the scene from the predicted layered representation, accounting for both occlusion and alpha blending.}
    \label{fig:rgbad-comp-hand}
    \vspace{-3mm}
\end{figure*}

\vspace{-2mm}
\section{Experiments and Results}
After evaluating our method quantitatively in Section~\ref{sec:experiments} we now present additional qualitative experiments and possible use cases of our method. We begin by using our predicted layered representations to re-composite images with additional geometry, first in a controlled setting using our renderer, then on increasingly difficult input data like screenshots and even scans of printed visualizations. After that we show that our representation allows for novel view synthesis through re-projection and that we can make use of the predicted depth map to create geometry-aware surface labels, as well as image based lighting. Lastly we highlight limitations we encountered with our method with a focus on the influence of illumination on our method.

\subsection{Re-composition with additional Geometry}
In this experiment, we first predict the layered representation of the input image. We use this layered representation to re-composite the input image with additional geometry. To composite the new geometry with the layered representation, we use the \emph{Depth} map together with the \emph{Back} to render intersections of the new geometry with the relevant structures, before rendering the \emph{Front} on top. When blending the \emph{Front} on top, the additionally predicted FirstHit depth is used to determine the \emph{Front}'s thickness. The results can be seen in Fig.~\ref{fig:rgbad-comp-hand}, where we compare our approach to naively rendering the additional geometry on top of the input image, as well as with the composited ground truth layers.

\noindent{\textbf{Discussion.}}
As can be seen in Fig.~\ref{fig:rgbad-comp-hand-naive}, the naive approach to compositing an RGB image with additional geometry usually does not produce any meaningful result, as the geometry simply occludes the image. Our approach on the other hand manages to correctly occlude the cylinders behind the bones of the fingers. Additionally, the cylinders are partially occluded behind the semi-transparent skin of the hand. Compared to the ground truth, our approach results in a slightly more opaque and darker skin color, but is visually still very close to the ground truth.

\subsection{Application in the Wild: Screenshots \& Prints}\label{sec:in-the-wild}
In the previous experiments we evaluated our approach on test datasets that are generally close to the training dataset, in the sense that most of the evaluated images contained either renderings of skulls from the CQ500 dataset that we excluded from training, or other human parts, e.g., the hand CT from Fig.~\ref{fig:rgbad-comp-hand}. To investigate how our approach generalizes beyond this data, we have applied it to renderings of a salamander (see Fig.~\ref{fig:rgbad-comp-nonhead2}). While these images still have been generated with the same volume renderer we have used to generate our training data, the ultimate test is to see how our method generalizes beyond such a controlled setting. Therefore, to evaluate such an applicability in the wild, we first run our method on screenshots taken from the seminal VolumeShop paper~\cite{bruckner2005volumeshop}, before going one step further, applying it to scans of a printed copy of the \emph{Symposium on Volume and Point-Based Graphics}~\cite{pajarola2008proceedings}. Those screenshots and scans partially contain labels and zoom ins, used different shading techniques and aspect ratios. We use a simple flood-fill algorithm for a very rough background extraction, and resize the images to the closest resolution divisible by $64$ (as required by our neural net), before predicting a layered representation. Finally, we re-composite the input image with additional geometry. Additionally we can use the predicted layers to compute ambient occlusion shading. The results of the screenshots and scans are shown in Fig.~\ref{fig:volumeshop} and Fig.~\ref{fig:wildprint} respectively. For the opaque objects, where the predicted depth map is close to the \emph{FirstHit} depth, we make a slight change to the compositing. If the object is mostly opaque, errors in the \emph{Front} layer can become visually quite obvious, when the \emph{Front} is blend over the additional geometry, where it should fully occlude the object. Consequently, we make use of the fact that our layered representation composites to an RGBA version of the input image. Using the original input RGB, the predicted alpha of the re-composition and the predicted depth map, we can perform standard depth-based alpha blending.

\noindent{\textbf{Discussion.}} The results for the salamander shown in Fig.~\ref{fig:rgbad-comp-nonhead2} indicate, that our approach learns the actual geometry rather than just the properties of volume rendered human skulls, as we can generate appealing results for other specimen. As shown in Fig.~\ref{fig:volumeshop} and~\ref{fig:wildprint}, our approach allows for compositing new geometry into existing volume renderings, by only using a single RGB image extracted as a screenshot or scan without having any access to the renderer or the underlying volume data. While we have no ground truth for these predictions, it is still quite clear that the quality of the depth maps still leaves room for improvement. For example in the depth prediction of the scanned skull (Fig.~\ref{fig:wildprint}, most of the blood vessels are not visible and some of the dark shadows in the split-open head (Fig.~\ref{fig:volumeshop}) cause holes in the predicted depth map. Despite those limitations our method already enables meaningful re-compositions of visualizations taken from the wild.

\subsection{Novel View Synthesis}
Using our predicted depth maps, we can re-project the scene into different cameras, enabling a viewpoint change. As our approach is currently limited to a two-layer representation, the viewpoint cannot change drastically, but enough to produce convincing stereo pairs for anaglyphs (compare Fig.~\ref{fig:anaglyph-chroma}), subtle animations or wiggle stereo (compare Supplemental Video), in order to improve the depth perception of the rendering.

\subsection{Surface Labels}
\begin{figure}[h]
    \centering
    \begin{subfigure}{0.4\linewidth}
    \includegraphics[width=\textwidth]{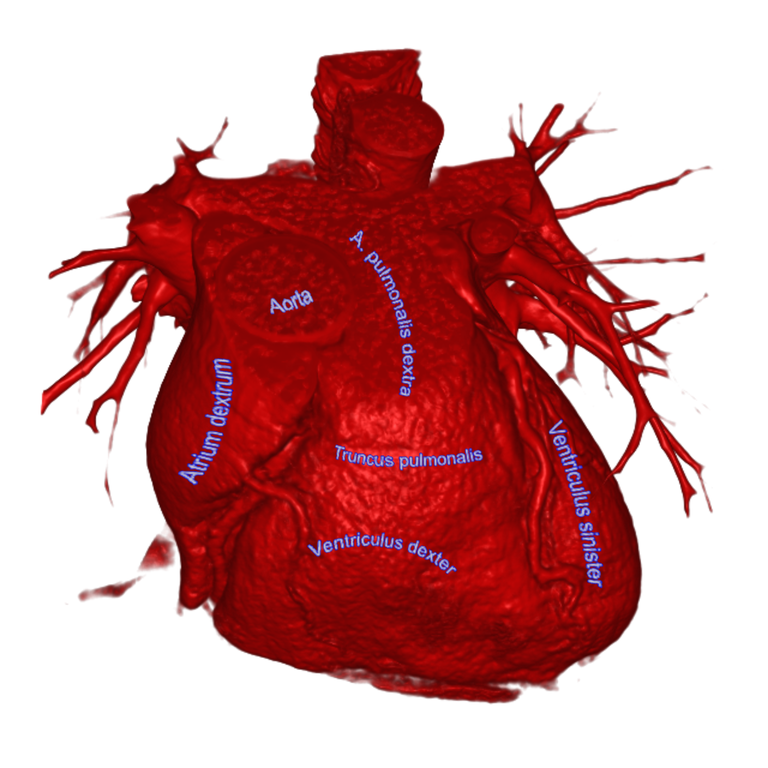}  
    \end{subfigure}%
    \begin{subfigure}{0.6\linewidth}
    \includegraphics[width=\textwidth]{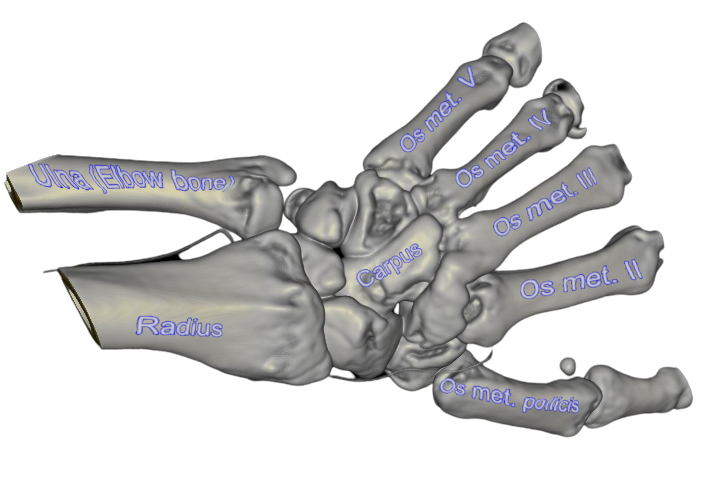}
    \end{subfigure}%
    \caption{\textbf{Surface Labeling.} Using our depth predictions we can add text to surfaces that respects the 3D shape of the objects, reproducing two Figures by Ropinski\etal~\cite{ropinski2007internal}.}
    \label{fig:surface_labels}
    \vspace{-1mm}
\end{figure}

Following the approach by Ropinski\etal~\cite{ropinski2007internal} we use the predicted depth to properly attach text labels to surfaces seen in the rendering. Respecting the surface geometry for text labels provides additional depth cues to the viewers, helping them to understand the geometry of the scene. To assess our method we tried to reproduce two Figures from their paper in Figure~\ref{fig:surface_labels}. Using our approach a medical illustrator can retrospectively add such labels to existing visualizations.

\begin{figure}[t]
    \caption{\textbf{Anaglyphs and Chromadepth Contours.} \emph{Left} is an anaglyph, to be viewed with blue-red stereo glasses. \emph{Right} shows the Chromadepth Fresnel effect by Behrendt\etal~\cite{behrendt2017combining}. Original rendering taken from the VolumeShop paper~\cite{bruckner2005volumeshop} (\textcopyright~2005~IEEE). Unfortunately we cannot show this Figure in the pre-print, please acquire our article from IEEE.}\label{fig:anaglyph-chroma}
    \vspace{-1mm}
\end{figure}

\begin{figure}[!h]
    \centering
    \begin{subfigure}{0.25\linewidth}
    \includegraphics[width=\textwidth]{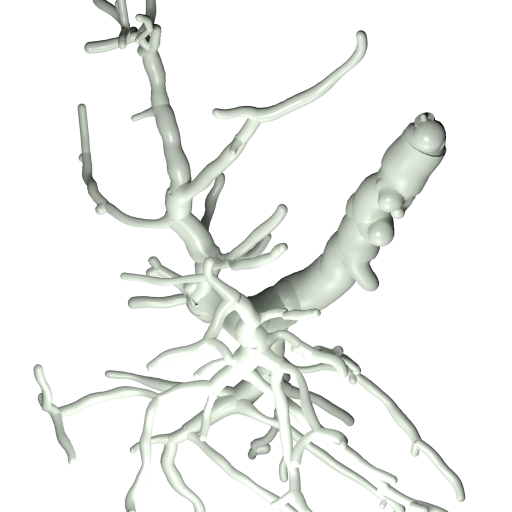}
    \caption{Input}
    \end{subfigure}%
    \begin{subfigure}{0.25\linewidth}
    \includegraphics[width=\textwidth]{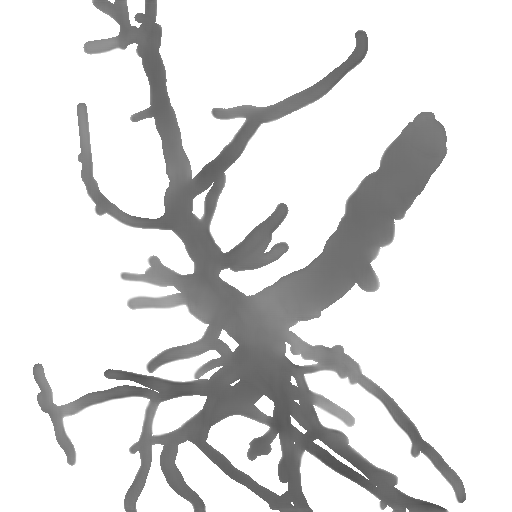}
    \caption{Depth}
    \end{subfigure}%
    \begin{subfigure}{0.25\linewidth}
    \includegraphics[width=\textwidth]{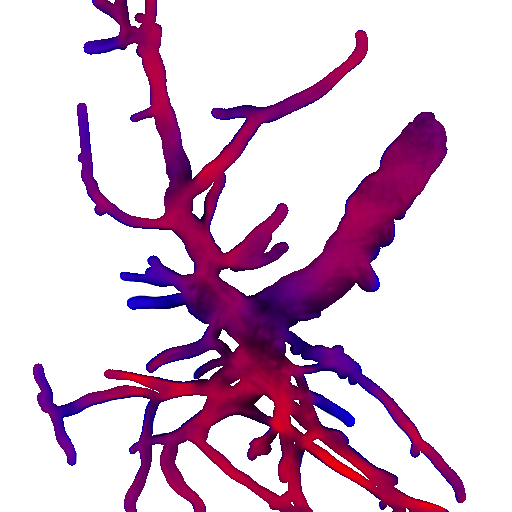}
    \caption{Chromadepth}
    \end{subfigure}%
    \begin{subfigure}{0.25\linewidth}
    \includegraphics[width=\textwidth]{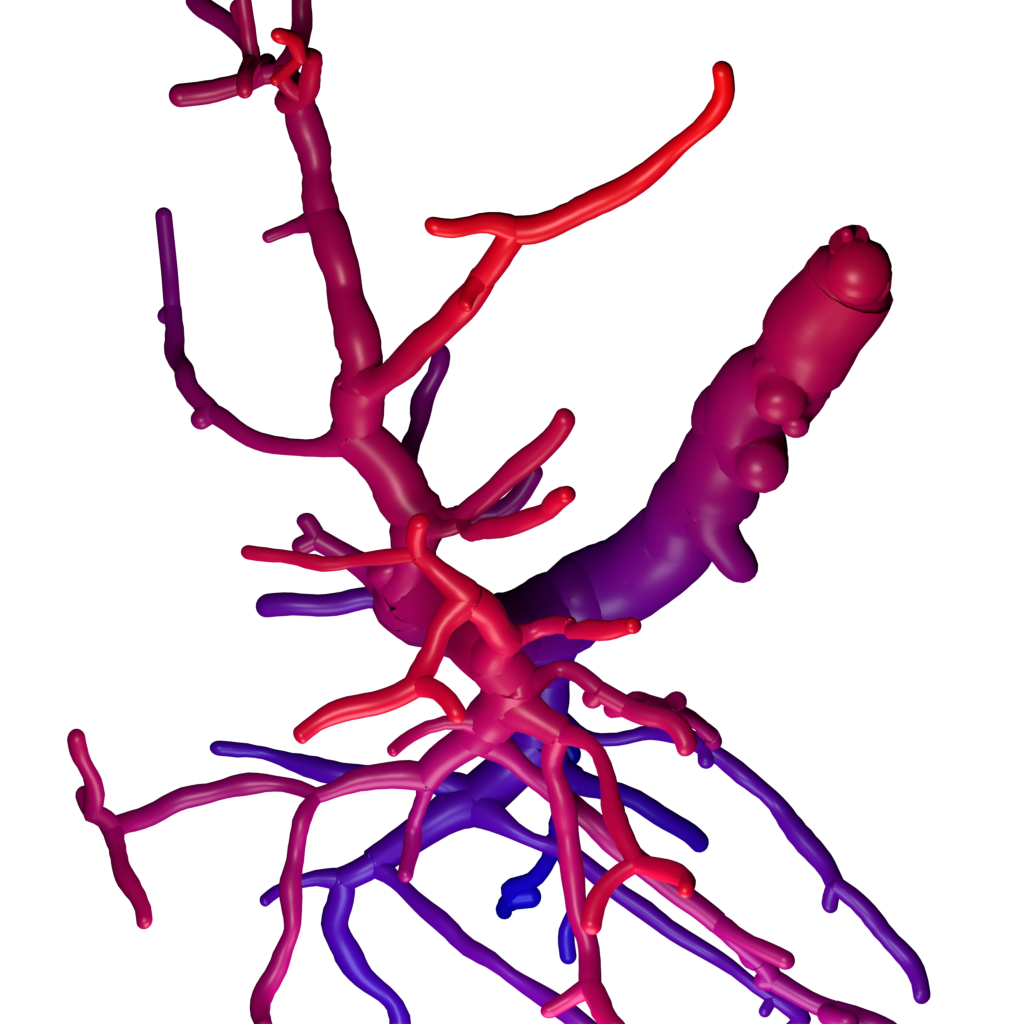}
    \caption{Target}
    \end{subfigure}%
    \caption{\textbf{Failure Case Angiography.} Our approach fails to correctly predict the depth on vessel structures. Many of the vessels are predicted to branch into the wrong direction. Even for humans it is notoriously difficult to correctly estimate depth in the presented vessel images. The two images on the right show the depth color coded onto the surface of the vessels (red is close, blue is far).}
    \label{fig:vessel-fail}
\end{figure}

\subsection{Image-Based Lighting}
Lastly we show that our approach can be used to add or modify the shading used in a visualization to some extent. In Figure~\ref{fig:rgbad-comp-nonhead2}~and~\ref{fig:volumeshop} we apply additional ambient occlusion to the renderings - also in combination with additional geometry, which helps to produce a convincing composition. We further use the predicted depths to estimate surface normals, enabling the application of contour shading and contour-based chromadepth shading~\cite{behrendt2017combining}, as illustrated in Fig.~\ref{fig:anaglyph-chroma}.

\subsection{Limitations}
While our approach enables a rich variety of modifications that we can apply to existing visualizations from just a single RGB image, an accurate depth prediction is essential to achieve convincing results. When applying our approach to images outside of our training domain, i.e. CT scans of human heads, we can see degrading depth quality the further we move away from this training domain. Here we discuss the limitations of our method and failure cases we discovered, especially when applied in the wild.

We see the first major degradation in Figure~\ref{fig:volumeshop} and \ref{fig:wildprint}, where you can see \emph{holes} in the depth maps where the input showed very deep black shadows. We suspect this is due to the different shading techniques used in those renderings, including contours and shadows, that deviate strongly from our training domain that only used Phong shading. It became clear to us that the issues especially in very dark regions emerge from the training data, that shows exclusively black backgrounds for which we train our model to predict a depth of $1$. This problem could possibly be resolved by including varying backgrounds during training.

Another limitation we found is that - especially out-of-domain images - result in poor separation of \emph{Front} and \emph{Back} layers. The resulting re-composition of those layers usually remains visually very similar to the input image, with both the \emph{Front} and \emph{Back} being a slightly transparent version of the input. While we introduce a special loss to combat this, you can still see this problem (see Fig.~\ref{fig:rgbad-ablation}). This still allows us to use the predicted depth map together with either the input image or the re-composited prediction that also includes an alpha channel, but lacks the separation of a semi-transparent \emph{Front}. We show further examples for this in our Appendix.

As a more challenging example in the medical domain we also tested our approach on angiography visualizations, as illustrated in Figure~\ref{fig:vessel-fail}. We hoped to retrospectively apply methods to improve the depth perception in such renderings, which is typically very hard for humans. The Figure shows an attempt to employ pseudo-chromadepth~\cite{steenblik1987chromostereoscopic, bailey1998using} shading to a vessel visualization, however our approach fails to correctly identify the depth of the different vessel branches and is thus not suitable for this task.

\begin{table}[tb]
    \centering
    \begin{tabular}{cllcccccc}
    & & \multicolumn{2}{c}{\includegraphics[width=1.6cm]{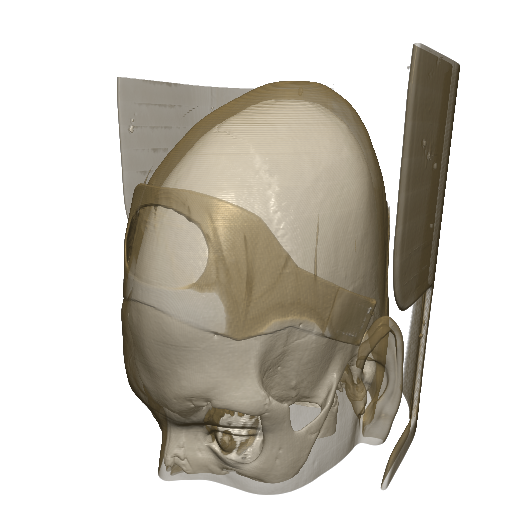}}
      & \multicolumn{2}{c}{\includegraphics[width=1.6cm]{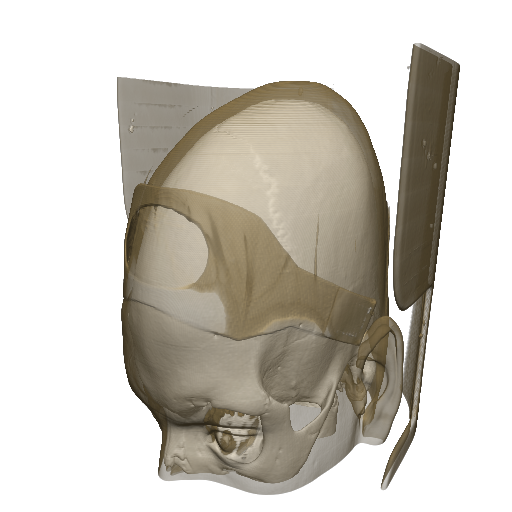}}
      & \multicolumn{2}{c}{\includegraphics[width=1.6cm]{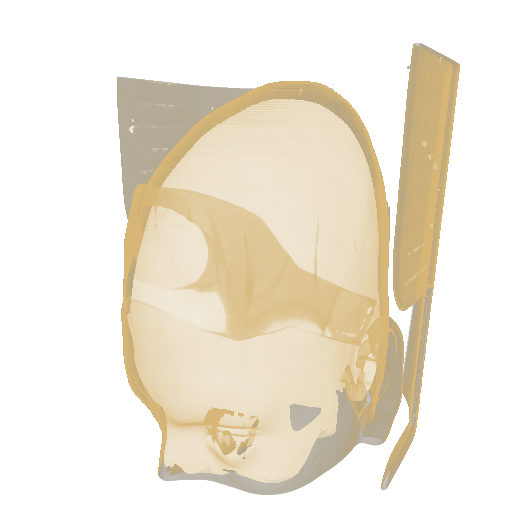}} \\
    \toprule
    & Predict on $\rightarrow$
    & \multicolumn{2}{c}{EA+Phong} & \multicolumn{2}{c}{EA+Diffuse} & \multicolumn{2}{c}{EA} \\
    \cmidrule(lr){3-4}
    \cmidrule(lr){5-6}
    \cmidrule(lr){7-8}
    & $\downarrow$ Trained on & $\bm\delta_1 \uparrow$ & $\bm{L}_1\downarrow$ & $\bm\delta_1 \uparrow$ & $\bm{L}_1\downarrow$ & $\bm\delta_1 \uparrow$ & $\bm{L}_1\downarrow$ \\
    \midrule
     \multirow{4}{*}{\rotatebox[origin=l]{90}{\scriptsize WYSIWYP}} 
     &  & &  &  & & &  \\[-2mm]
       & EA+Phong   & .970 & .018 & .967 & .017 & .953 & .021 \\
       & EA+Diffuse & \textbf{.971} & \textbf{.016} &\textbf{.971} & .017 & .954 & .021 \\
       & EA         & .965 & .019 & .965 & .019 & .966 & .020 \\
       &  & &  &  & & &  \\[-2mm]
    \midrule
     &  & &  &  & & &  \\[-2.5mm]
    \multirow{3}{*}{\rotatebox{90}{\scriptsize FirstHit}}
       & EA+Phong   & .986 & .012 & .986 & .011 & .976 & .015 \\
       & EA+Diffuse & \textbf{.989} & \textbf{.010} & .988 & \textbf{.010} & .979 & .014 \\
       & EA         & .986 & .013 & .986 & .013 & .985 & .013 \\
    \bottomrule
    \end{tabular}
    \caption{\textbf{Influence of Illumination.} We compare models trained (and predicting) on three datasets using different illumination techniques, namely standard \emph{Emission-Absorption} (EA) without shading, with a diffuse component and with the full Phong shading model. The diffuse component appears to be most relevant for estimating depth. When training or predicting on EA without shading, the model and predictions become significantly worse.}
    \label{tab:illumination-ablation}
    \vspace{-1.2mm}
\end{table}

\noindent{\textbf{Influence of Illumination.}}
As the illumination in a volume rendering has a great influence on human depth perception, we checked if this is also true for our neural network. We then found that our model performs worse on renderings that only use a simple emission-absorption model without additional diffuse or specular shading. To further investigate this we performed an additional experiment (compare Table~\ref{tab:illumination-ablation}).
We re-trained our method on datasets with different shading techniques: (1) emission-absorption (EA) only (i.e. merely applying the TF without shading), (2) EA with diffuse shading, and (3) using the full Phong shading model including specular highlights. Already being aware that the naive background strategy in our original datasets may cause problems, we also generated all those datasets with randomized backgrounds that were either black or white and contained shot noise (details in the Appendix). This also lead to a slight improvement over our numbers reported in Table~\ref{tab:rgbad-ablation}.

\noindent{\textbf{Discussion.}} We can see that, while there is a significant decrease in depth estimation performance for the EA only case, none of the models completely fails on differently lit renderings. However, including at least diffuse shading helps the model significantly, but at the same time the models that were not trained on EA-only renderings, see a drop in performance when predicting EA-only images. We would suspect, that models trained on shaded datasets generalize better to real-world renderings, as they usually contain some form of shading, but including a mix of illumination techniques in the training dataset could help to make the model more invariant towards illumination in general. 


\section{Conclusions and Future Work}\label{sec:conclusion}
Within this paper, we have presented the first systematic study investigating the capabilities of deep monocular depth estimation techniques in the context of volume rendered images containing semi-transparent structures. We could show, that SotA monocular depth estimators can be used to predict meaningful depth maps for such images. Motivated by these findings, we have extended the \textsc{BTS} approach, to not only allow for the estimation of depth, but also of a layered image representation containing color and opacity. Our extension not only modulates the output of the model, but also facilitates residual learning and exploits a novel compositing loss and regularization. With the presented approach, we are able to generate meaningful layered representations of RGB-only volume rendered images, and are thus able to modify them according to visualization-centric tasks, such as for instance the integration of mesh geometry, surface labels or application of image based lighting.

In the future, we would like to investigate how we can increase the amount of layers predicted. As prior work on layered depth images has shown, most methods operating on such a layered representation greatly benefit from more layers. More layers could enable novel view synthesis beyond the subtle changes we achieve, more complex lighting and possibly even changing the transfer function, just to name a few opportunities. We would also like to systematically evaluate the influence of the underlying shading model on the depth estimation and layer decomposition, including illustrative shading techniques like contours and silhouettes.

\begin{figure*}[!h!t]
    \centering
    \caption{\textbf{Applying our approach in the wild.} Here we took screenshots from the seminal VolumeShop paper~\cite{bruckner2005volumeshop} (\textcopyright~2005~IEEE). Unfortunately we cannot show this Figure in the pre-print, please acquire our article from IEEE.}
    \label{fig:volumeshop}
\end{figure*}

\begin{figure*}[h!]
    \centering
    \caption{\textbf{Applying our approach in the wild.} Here we took scans from a printed copy of the \emph{Symposium on Volume and Point-Based Graphics}~\cite{pajarola2008proceedings} (\textcopyright~2008~IEEE). Unfortunately we cannot show this Figure in the pre-print, please acquire our article from IEEE.}
    \label{fig:wildprint}
\end{figure*}

\vspace{-1mm}
\section*{Acknowledgments}
This work was partially funded by the Deutsche Forschungsgemeinschaft (DFG) under grant 391107954 (Inviwo). Some of the figures were produced using the Inviwo framework (\url{https://inviwo.org}).
\bibliographystyle{abbrv-doi}
\vspace{-1mm}
\bibliography{template}
\vspace{3mm}
\begin{IEEEbiography}[{\includegraphics[width=1in,height=1.25in,clip,keepaspectratio]{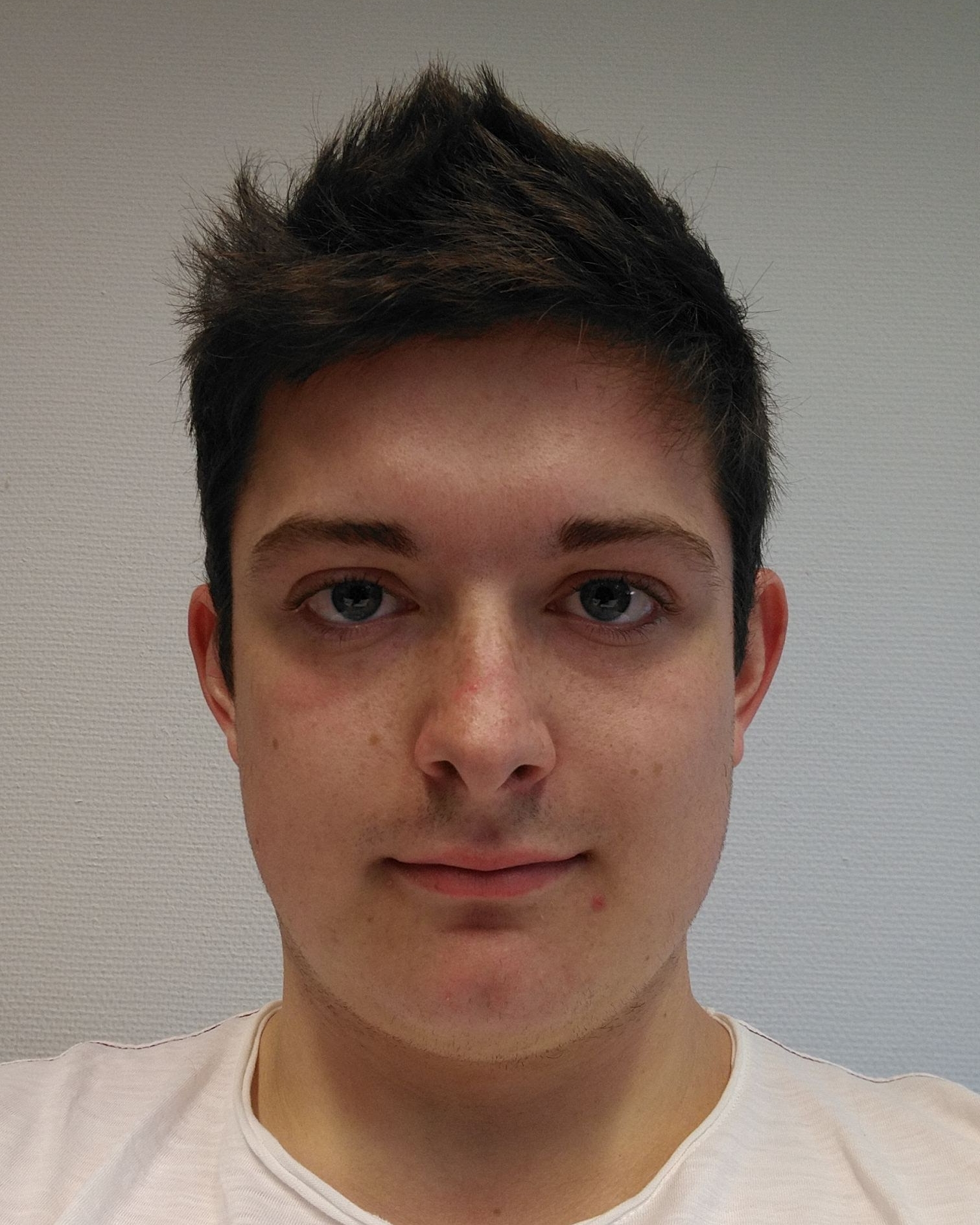}}]{Dominik~Engel}
is a Ph.D. student at Ulm University, Germany, where he previously received his B.Sc. and M.Sc. degrees in computer science. In 2018 he joined the Visual Computing research group. His research focuses on deep learning in visualization and computer graphics, differentiable and neural rendering.
\end{IEEEbiography}
\vspace{-10mm}
\begin{IEEEbiography}[{\includegraphics[width=1in,height=1.25in,clip,keepaspectratio]{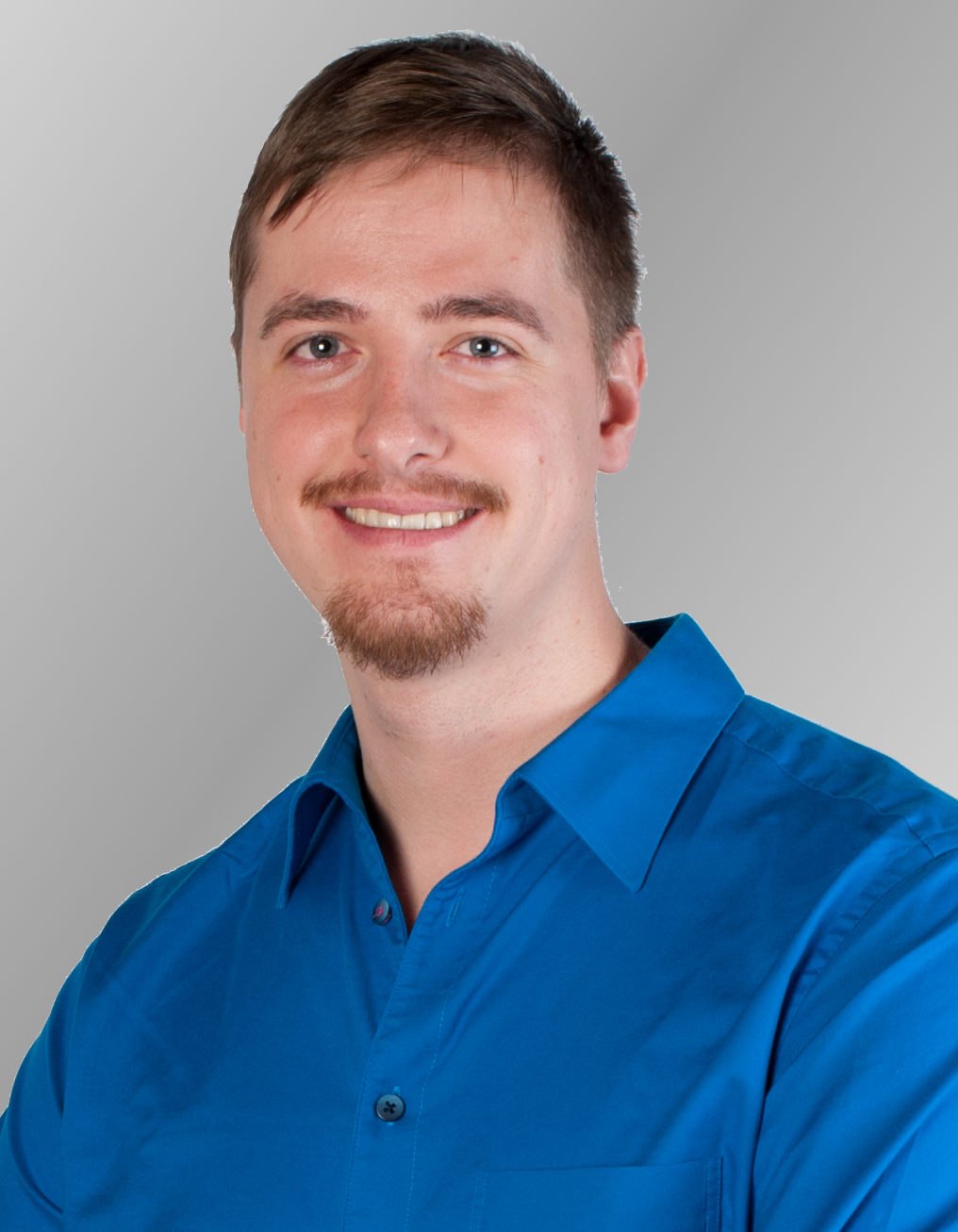}}]{Sebastian~Hartwig}
is a Ph.D. student at Ulm University, Germany, where he previously received his B.Sc. and M.Sc. degrees in computer science. He completed his masters degree at the Institute of Media Informatics 2017 before joining the research group Visual Computing. His research focus on machine scene understanding like depth and layout estimation, as well as photorealistic rendering for synthesizing high quality training data.
\end{IEEEbiography}
\vspace{-10mm}
\begin{IEEEbiography}[{\includegraphics[width=1in,height=1.25in,clip,keepaspectratio]{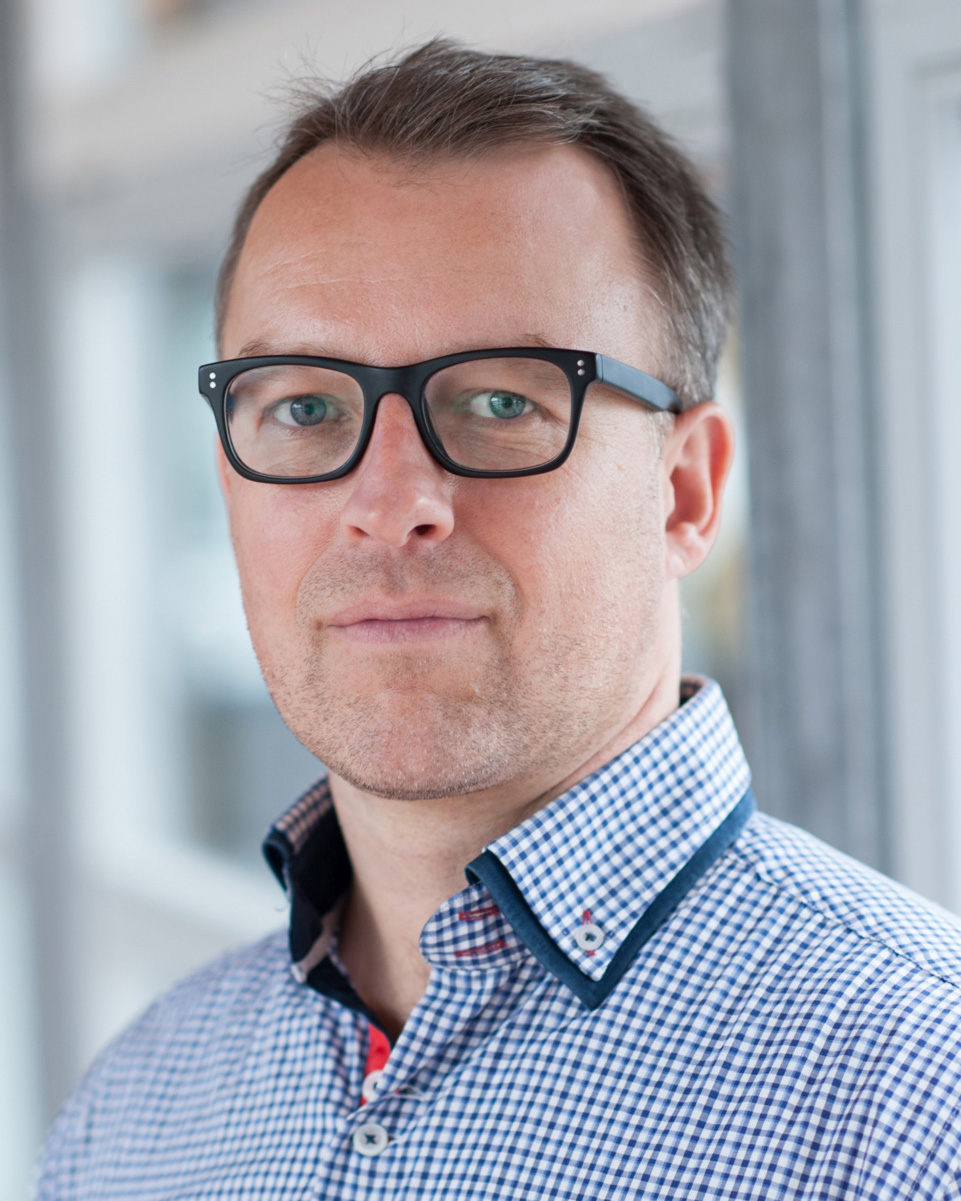}}]{Timo~Ropinski}
is a professor at Ulm University, heading the Visual Computing Group. Before moving to Ulm, he was Professor in Interactive Visualization at Linköping University, heading the Scientific Visualization Group. He received his Ph.D. in computer science in 2004 from the University of Münster, where he also completed his Habilitation in 2009. Currently, Timo serves as chair of the EG VCBM Steering Committee, and as an editorial board member of IEEE TVCG.
\end{IEEEbiography}

\newpage
\appendices

\section{Comparison of Depth Techniques}
\begin{figure}[h!]
    \centering
    \begin{subfigure}{0.5\linewidth}
        \begin{subfigure}{0.5\textwidth}
        \includegraphics[width=\textwidth]{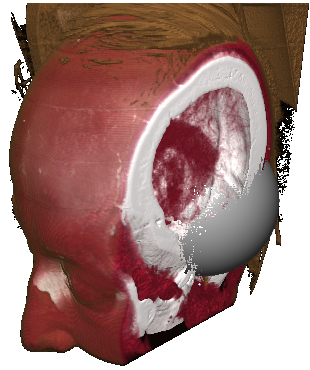}
        \\
        \includegraphics[width=\textwidth]{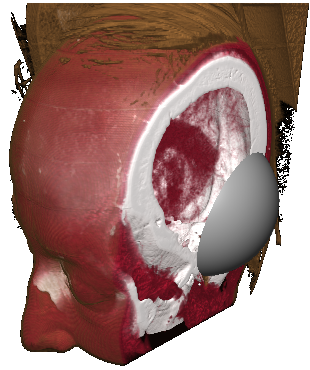}
        \end{subfigure}%
        \begin{subfigure}{0.5\textwidth}
        \includegraphics[width=\textwidth]{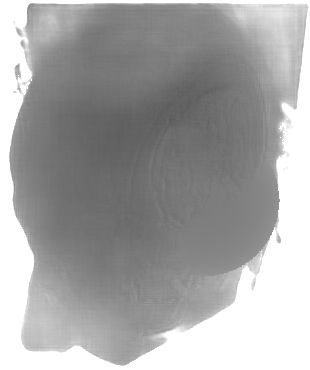}
        \\
        \includegraphics[width=\textwidth]{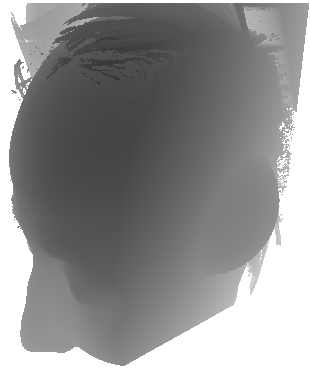}
        \end{subfigure}%
        \caption{First Hit}
    \end{subfigure}%
    \begin{subfigure}{0.5\linewidth}
        \begin{subfigure}{0.5\textwidth}
        \includegraphics[width=\textwidth]{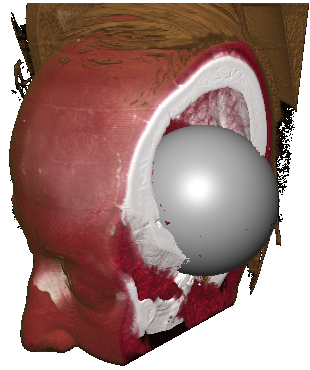}
        \\
        \includegraphics[width=\textwidth]{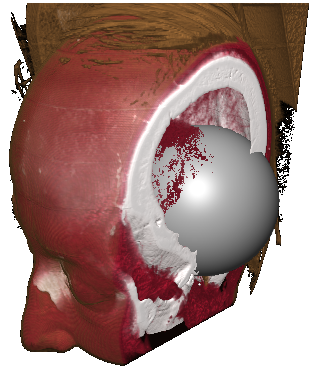}
        \end{subfigure}%
        \begin{subfigure}{0.5\textwidth}
        \includegraphics[width=\textwidth]{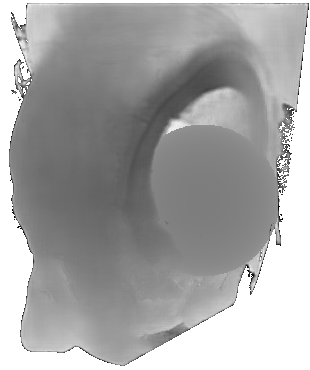}
        \\
        \includegraphics[width=\textwidth]{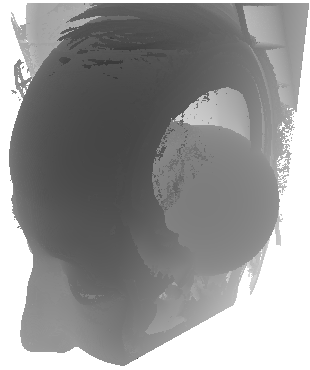}
        \end{subfigure}%
        \caption{WYSIWYP}
    \end{subfigure}%
    \caption{\textbf{Comparison First Hit against WYSIWYP for compositing}. Predictions on \emph{top}, targets on \emph{bottom}, each showing the resulting composition on the \emph{left} and the depth on the \emph{right}. Note how \emph{First Hit} depth causes occlusion of the sphere inside the skull, while techniques that focus on capturing the relevant geometry, like WYSIWYP, allow for a convincing composition.}
    \label{fig:composite-depth-fh-vs-wys}
\end{figure}

As shown in the main paper, the different presented \emph{depth techniques} have a large impact on the resulting depth map of a semi-transparent rendering. As one of our presented applications is the composition with additional objects, we want to highlight that such composition can be heavily influenced by the choice of depth technique. Figure~\ref{fig:composite-depth-fh-vs-wys} illustrates this effect by comparing composition using FirstHit depth to using WYSIWYP depth. This experiment does not consider blending of colors and alpha, but only shows the color with lower depth for each pixel. When compositing semi-transparent scenes we expect the additional geometry to be placed in accordance with the visually relevant surfaces of the scene, like the backside of the skull in the illustration. Using the FirstHit depth, the highly - but not fully - transparent interior of the skull \emph{occludes} the sphere during composition, disallowing a meaningful composition with the scene.

\begin{figure}[h!]
    \centering
    \newcommand{\fidx}{2dc}
    \begin{subfigure}{0.33\linewidth}
        \includegraphics[width=\textwidth]{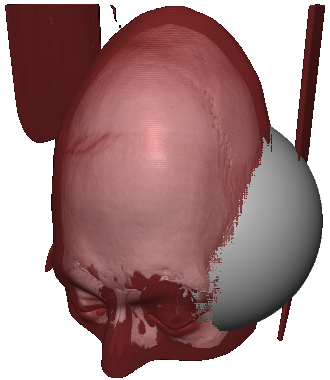}
        \\
        \includegraphics[width=\textwidth]{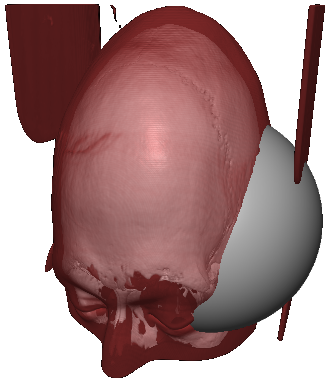}
        \caption{Composited}
    \end{subfigure}%
    \begin{subfigure}{0.33\linewidth}
        \includegraphics[width=\textwidth]{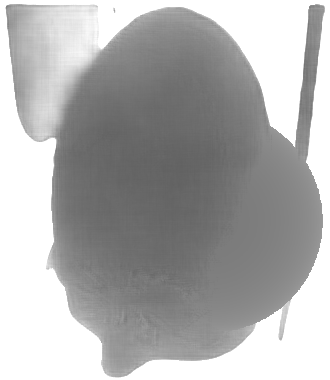}
        \\
        \includegraphics[width=\textwidth]{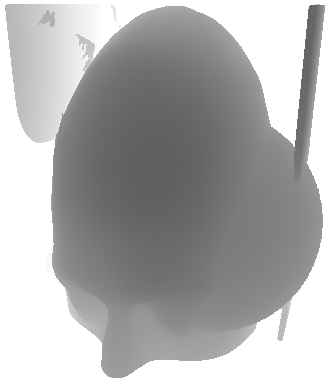}
        \caption{Depth}
    \end{subfigure}%
    \begin{subfigure}{0.33\linewidth}
        \includegraphics[width=\textwidth]{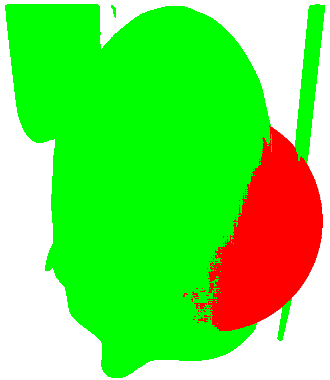}
        \\
        \includegraphics[width=\textwidth]{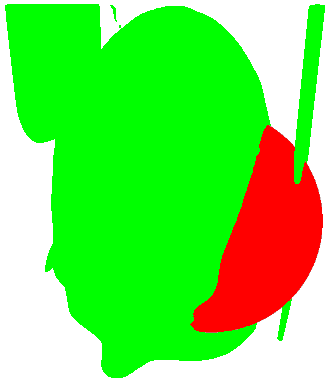}
        \caption{In Front}
    \end{subfigure}%
    \caption{\textbf{Depth-only Compositing} of the input image with an additional object, using the predicted depth map and the input image. The \emph{top} row shows predictions, \emph{bottom} row shows ground truth. The right image (\emph{c}) shows which of the objects is in front.}
    \label{fig:composite-depth}
\end{figure}

\section{Input pre-processing}
When making predictions on new renderings we require a simple pre-processing step in order to get a rough background segmentation and to comply with the neural net's restrictions on the input resolution. While our examples in Fig.~\ref{fig:volumeshop} resize the input images to $512\times 512$ pixels, our approach is not limited to this resolution. Generally the BTS network is fully convolutional and thus not bound to a fixed input size. One however needs to supply images with widths and heights that are divisible by $64$, otherwise the encoder and decoder of the network would have mis-matching feature tensor sizes due to the pooling operations, preventing the use of skip connections. In practice a given input image can be resized to the closest resolution that is divisible by $64$ for minimal distortion.

\begin{table*}[h!tb]
  \centering
  \begin{tabular}{lccccccccc}
    \toprule
    \textsc{\textbf{Layer}} & \textsc{Metric} & \textsc{Base} & \textsc{+R} & \textsc{+C} & \textsc{+FB} & \textsc{+R+C} & \textsc{+R+FB} & \textsc{+C+FB} & \textsc{+R+C+FB} \\
    \midrule
    Front & \multirow{3}{*}{SSIM $\uparrow$}
                &  .731 & .744 & .715 & .718 & .800 & .780 & .706 & \textbf{.802} \\
    Back       &&  .689 & .827 & .725 & .798 & .798 & .829 & .664 & \textbf{.833} \\
    Composite  &&  .830 & .894 & .822 & .838 & .958 & .930 & .792 & \textbf{.961} \\

    \rule{0pt}{1mm} &&&&& \\ 
    Front & \multirow{3}{*}{$\bm{L}_1\downarrow$}
                & .070 & .073 & .068 & .068 & .073 & .071 & \textbf{.066} & .073 \\
    Back       && .080 & .087 & .075 & .078 & .081 & .083 & \textbf{.073} & .079 \\
    Composite  && .043 & .046 & .030 & .041 & \textbf{.025} & .043 & .029 & \textbf{.025} \\

\rule{0pt}{1mm} &&&&& \\ 
    \textsc{\textbf{Depth}} && \multicolumn{4}{c}{} \\
    \midrule
    FirstHit & \multirow{2}{*}{$\bm\delta_1\uparrow$}
              & \textbf{.981} & .980 & .980 & \textbf{.981} & \textbf{.981} & .\textbf{981} & \textbf{.981} &\textbf{.981} \\
    WYSIWYP  && .964 & .963 & .962 & .964 & \textbf{.966} & .964 & .964 & .964 \\

    \rule{0pt}{1mm} &&&&& \\ 
    FirstHit & \multirow{2}{*}{$\bm{L}_1\downarrow$}
              & .015 & .015 & .016 & .015 & \textbf{.014} & .015 & .015 & \textbf{.014} \\
    WYSIWYP  && .021 & .022 & .022 & .021 & \textbf{.019} & .021 & .021 & .021 \\
    \bottomrule
  \end{tabular}
  \caption{\textbf{Ablation study} of the \emph{Residual Image Output} (+R), \emph{Compositing Loss} (+C) and \emph{Front-Back-Divergence} (+FB).}
  \label{tab:rgbad-ablation}
\end{table*}

\begin{figure*}[h!tb]
    \caption{\textbf{Additional results and pipeline.} Here we took screenshots (\emph{a}) from the seminal VolumeShop paper~\cite{bruckner2005volumeshop} (\textcopyright~2005~IEEE). Unfortunately we cannot show this Figure in the pre-print, please acquire our article from IEEE.}
    \label{fig:volumeshop}
\end{figure*}

\section{Additional Ablations}
In the main paper we present an ablation study of our proposed network modifications, namely \emph{Residual Image Outputs} (+R), \emph{Compositing Loss} (+C) and \emph{Front-Back-Divergence} (+FB). The results reported in the main paper focus on the build up of these three techniques and how they resolve the main issues encountered during development of our approach. Here we report all 8 combinations in Table~\ref{tab:rgbad-ablation}.

\section{Randomizing Backgrounds}
For randomizing the backgrounds of our training images in the \emph{Influence of Illumination} experiment, we decided to use a white-ish or black-ish background with a 50\% chance. Hereby white-ish and black-ish mean gray values uniformly sampled from the ranges $[0.95, 1.0]$ and $[0.0, 0.05]$ respectively. In addition to that we added salt and pepper noise. Specifically we set 200 randomly selected pixels of the background to $1.0$ and 200 other random pixels to $0.0$.

\section{Additional Failure Cases}
We present additional examples where our approach fails to make good predictions. As stated in the main paper, our approach degrades with renderings that deviate strongly from our training domain. For example both the hand and engine in Figure~\ref{fig:volumeshop} show significant problems. One can see that for the hand example, our approach mostly manages to separate the single blood vessel on the right, while the depth of the remaining hand is predicted to be mostly flat. While we have no ground truth to compare this in detail, it is clear that the vessel on the left, the elbow bone on the bottom and the individual fingers should be recognizable in the depth map. While the engine predictions reveal more pronounced structures, many of the fine details are missing in the depth maps and the overall proportions are incorrect.

\begin{figure*}
    \centering
    \begin{subfigure}{0.199\textwidth}
        \includegraphics[width=\textwidth]{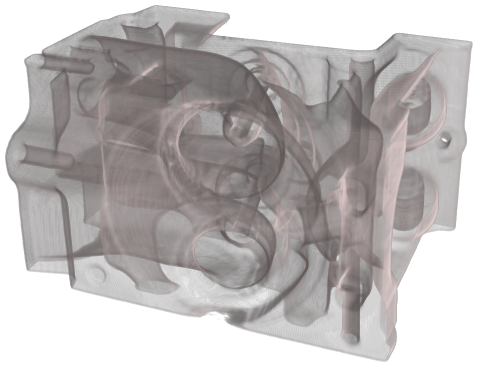}
        \caption{Input}
    \end{subfigure}%
    \begin{subfigure}{0.199\textwidth}
        \includegraphics[width=\textwidth]{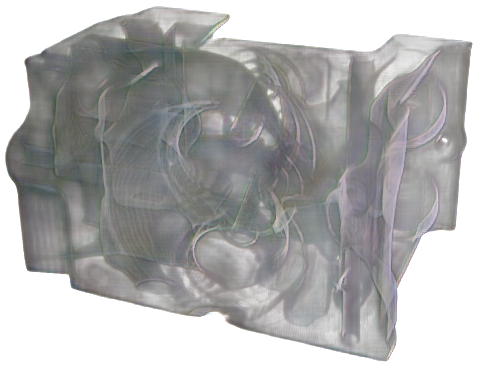}
        \caption{Re-Composited}
    \end{subfigure}
    \begin{subfigure}{0.199\textwidth}
        \includegraphics[width=\textwidth]{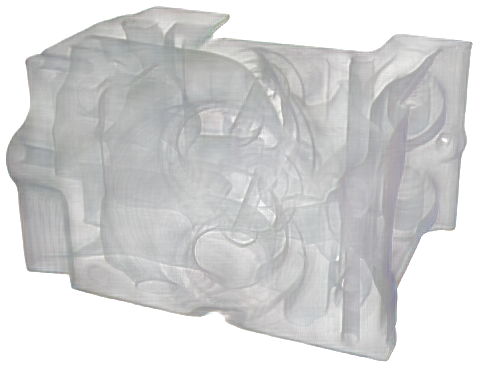}
        \caption{Front}
    \end{subfigure}%
    \begin{subfigure}{0.199\textwidth}
        \includegraphics[width=\textwidth]{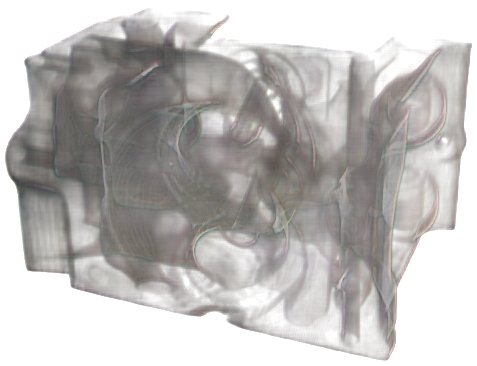}
        \caption{Back}
    \end{subfigure}%
    \begin{subfigure}{0.199\textwidth}
        \includegraphics[width=\textwidth]{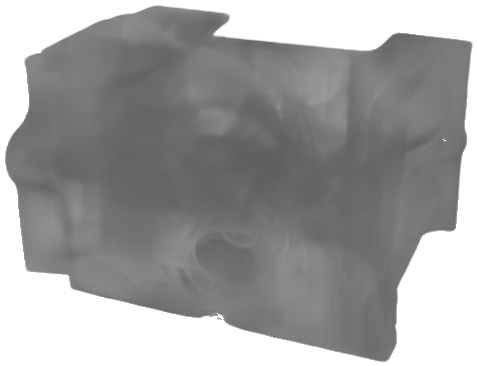}
        \caption{Depth}
    \end{subfigure}
    \caption{\textbf{Failure Case Engine.} The \emph{Front} and \emph{Back} are not properly separated. Both layers show a quite transparent version of the input image and the depth predictions fails to represent the detailed interior of the engine.}
    \label{fig:fail-engine}
\end{figure*}

Even more difficult examples, as shown in Figure~\ref{fig:fail-engine} lead to unusable depth maps and very indecisive front-back separation, i.e. both layers show basically the same blurry structure, so that it composites roughly to the input image. 
Similarly Figure~\ref{fig:fail-heart} shows again that deep black shadows cause holes in our model's predictions, not only in the depth map, but also in the alpha channels of the \emph{Front} and \emph{Back}.

\section{Full size Figures}
Lastly we show an additional full-size version of the stereo anaglyphs from the main paper, together with another anaglyph example in Figure~\ref{fig:big-anaglyphs}.

\begin{figure*}
    \caption{\textbf{Failure Case Heart.} This example again shows that deep black shadows lead to holes in the depth map and in this case also in the \emph{Front} and \emph{Back} predictions. Image taken as Screenshot from Lindemann\etal~\cite{lindemann2011influence} (\textcopyright~2011~IEEE). Unfortunately we cannot show this Figure in the pre-print, please acquire our article from IEEE.}
    \label{fig:fail-heart}
\end{figure*}

\begin{figure*}
    \caption{\textbf{Stereo Anaglyphs.} To be viewed with blue-red stereo glasses. The input images for this are again taken as screenshots from the VolumeShop paper~\cite{bruckner2005volumeshop} (\textcopyright~2005~IEEE). Unfortunately we cannot show this Figure in the pre-print, please acquire our article from IEEE.}
    \label{fig:big-anaglyphs}
\end{figure*}

\end{document}